\documentclass[sn-mathphys,Numbered]{sn-jnl}% Math and Physical Sciences Reference Style
\usepackage{graphicx}%
\usepackage{multirow}%
\usepackage{amsmath,amssymb,amsfonts}%
\usepackage{amsthm}%
\usepackage{mathrsfs}%
\usepackage[title]{appendix}%
\usepackage{xcolor}%
\usepackage{textcomp}%
\usepackage{manyfoot}%
\usepackage{booktabs}%
\usepackage{algorithm}%
\usepackage{algorithmicx}%
\usepackage{algpseudocode}%
\usepackage{listings}%
\DeclareMathOperator*{\argmin}{arg\,min}
\theoremstyle{thmstyleone}%
\theoremstyle{thmstyletwo}%

\theoremstyle{thmstylethree}%

\begin{document}

\title[Article Title]{Adaptive Activation Functions for Predictive Modeling with Sparse Experimental Data}

%%=============================================================%%
%% Prefix	-> \pfx{Dr}
%% GivenName	-> \fnm{Joergen W.}
%% Particle	-> \spfx{van der} -> surname prefix
%% FamilyName	-> \sur{Ploeg}
%% Suffix	-> \sfx{IV}
%% NatureName	-> \tanm{Poet Laureate} -> Title after name
%% Degrees	-> \dgr{MSc, PhD}
%% \author*[1,2]{\pfx{Dr} \fnm{Joergen W.} \spfx{van der} \sur{Ploeg} \sfx{IV} \tanm{Poet Laureate} 
%%                 \dgr{MSc, PhD}}\email{iauthor@gmail.com}
%%=============================================================%%

\author*[1]{\fnm{Farhad} \sur{Pourkamali-Anaraki}}\email{farhad.pourkamali@ucdenver.edu}

\author[2]{\fnm{Tahamina} \sur{Nasrin}}\email{tahamina\_nasrin@student.uml.edu}

\author[3]{\fnm{Robert E.} \sur{Jensen}}\email{robert.e.jensen.civ@army.mil}

\author[2]{\fnm{Amy M.} \sur{Peterson}}\email{amy\_peterson@uml.edu}

\author[4]{\fnm{Christopher J.} \sur{Hansen}}\email{christopher\_hansen@uml.edu}

\affil*[1]{\orgdiv{Department of Mathematical and Statistical Sciences}, \orgname{University of Colorado Denver}, \orgaddress{\street{1201 Larimer St}, \city{Denver}, \postcode{80204}, \state{CO}, \country{USA}}}

\affil[2]{\orgdiv{Department of Plastics Engineering}, \orgname{University of Massachusetts Lowell}, \orgaddress{\street{1 University Ave}, \city{Lowell}, \postcode{01854}, \state{MA}, \country{USA}}}

\affil[3]{\orgname{DEVCOM Army Research Laboratory}, \orgaddress{\city{Aberdeen Proving Ground}, \postcode{21005}, \state{MD}, \country{USA}}}

\affil[4]{\orgdiv{Department of Mechanical and Industrial Engineering}, \orgname{University of Massachusetts Lowell}, \orgaddress{\street{1 University Ave}, \city{Lowell}, \postcode{01854}, \state{MA}, \country{USA}}}

%%==================================%%
%% sample for unstructured abstract %%
%%==================================%%

\abstract{A pivotal aspect in the design of neural networks lies in selecting activation functions, crucial for introducing nonlinear structures that capture intricate input-output patterns. While the effectiveness of adaptive or trainable activation functions has been studied in domains with ample data, like image classification problems, significant gaps persist in understanding their influence on classification accuracy and predictive uncertainty in settings characterized by limited data availability. This research aims to address these gaps by investigating the use of two types of adaptive activation functions. These functions incorporate shared and individual trainable parameters per hidden layer and are examined in three testbeds derived from additive manufacturing problems containing fewer than one hundred training instances. Our investigation reveals that adaptive activation functions, such as Exponential Linear Unit (ELU) and Softplus, with individual trainable parameters, result in accurate and confident prediction models that outperform fixed-shape activation functions and the less flexible method of using identical trainable activation functions in a hidden layer. Therefore, this work presents an elegant way of facilitating the design of adaptive neural networks in scientific and engineering problems.}

\keywords{Predictive modeling, Neural networks, Activation functions, Conformal prediction, Small data}

%%\pacs[JEL Classification]{D8, H51}

%%\pacs[MSC Classification]{35A01, 65L10, 65L12, 65L20, 65L70}

\maketitle

\section{Introduction}\label{sec:intro}
Neural networks have made impressive strides in predictive modeling tasks due to their ability to learn nested or composite functions \cite{lu2020universal,talaei2023deep}. Therefore, the resulting expressive power to discern complex input-output relationships makes them suitable for a wide range of scientific and engineering disciplines, including biomedical engineering \cite{abdou2022literature,weiss2022applications,liu2023development}, structural engineering \cite{pourkamali2021neural,khodadadi2023non}, mechanical engineering \cite{olivier2021bayesian,stuckner2021optimal,brunton2020special,erichson2020shallow}, and additive manufacturing \cite{johnson2020invited,pourkamali2023evaluation}. Neural networks are made up of neurons that are interconnected units that process and transmit information. They do this by performing two pivotal operations: computing a weighted sum of inputs and then applying a nonlinear \textit{activation function}. This interplay among units equips neural networks to learn hierarchical features and multilayered representations to navigate the inherent complexities of diverse problems. In particular, neural networks excel in tasks that require simultaneous feature extraction and predictive modeling within high-dimensional feature spaces.

Among the various hyperparameters that users must set in advance, the selection of activation functions is a key element because of their role in encapsulating nonlinear patterns in the data. Typically, activation functions are predetermined and remain constant throughout the training process in the majority of scenarios. Classical examples of fixed activation functions include Sigmoid and hyperbolic tangent (tanh) functions \cite{hayou2019impact}, which squash input values into a specific range. However, fixed activation functions may suffer from problems such as the vanishing gradient problem \cite{hu2021handling}, where gradients become extremely small during backpropagation, hindering the learning process. Therefore, a wide range of activation functions with varying forms of nonlinearity have been introduced in recent years, including Rectified Linear Unit (ReLU) \cite{shen2022enhancement}, Exponential Linear Unit (ELU) \cite{clevert2015fast}, Softplus \cite{zheng2015improving}, and Swish \cite{ramachandran2017searching}, to name a few. For instance, the widely-used deep learning library Keras \cite{chollet2021deep}, as of version 2.14.0, offers a collection of 17 built-in activation functions.

Although the rapid increase in the number of fixed-shape activation functions offers potential benefits, a significant challenge arises from the intensive computational demands of conducting an exhaustive search to identify optimal functions for specific tasks. This challenge becomes more pronounced in dynamic and diverse scientific problems, necessitating frequent restarts of the search process to maintain the relevance of chosen activation functions. Consequently, the inflexibility inherent in fixed activation functions poses a hindrance to the streamlined development of neural network models within scientific and engineering disciplines.

To address this problem, one promising area of research is the use of adaptive or trainable activation functions \cite{agostinelli2014learning,lee2022stochastic,dubey2022activation}. Unlike traditional fixed activation functions such as ReLU or Sigmoid, which maintain a static form throughout training, adaptive functions evolve in response to the data distribution and the model's learning progress. These functions contain training parameters themselves, allowing us to optimize and tailor the shape or behavior of activation functions during the learning process alongside the parameters of the neural network. Therefore, this adaptability has the potential to enhance the ability to learn complex representations and patterns, while reducing the reliance on costly search procedures and detailed domain expertise.

Although the effectiveness of adaptive activation functions has been extensively studied in fields with abundant annotated data, such as image classification tasks highlighted in Tables 2 and 3 of a recent survey \cite{apicella2021survey}, a significant gap exists in understanding their applicability in domains characterized by sparse labeled data sets. A key concern in such scenarios, where data availability is limited, is the potential adverse impact on performance due to the increase in trainable parameters. Furthermore, current research has primarily relied on the standard classification accuracy score to evaluate the success of adaptive activation functions. This score, defined as the ratio of correct predictions to the total number of test samples, poses limitations when applied to scientific and engineering problems with a small number of samples. This metric offers limited assurance regarding the confidence and stability in neural network predictions. Hence, it is imperative to move towards metrics that quantify the uncertainty in the predictions of neural networks to provide a more nuanced understanding of the utility of adaptive activation functions in scientific problems.

Therefore, this paper aims to address the shortcomings mentioned above by systematically investigating the use of adaptive activation functions in three distinct additive manufacturing problems, each limited to fewer than 100 training samples. Our goal is to provide new insights into the possible application of adaptive activation functions, which introduce additional training parameters into data-austere environments. This study involves a comprehensive evaluation of the selection and adaptability of activation functions, coupled with an analysis of their impact on the predictive accuracy and confidence. In particular, we summarize our main contributions as follows.
\begin{enumerate}
    \item We study the effectiveness of adaptive activation functions compared to their counterparts using fixed-shape activation functions, namely ELU, Softplus and Swish. In our investigation, we explore the common practice of sharing activation functions within a hidden layer and the less explored terrain of using activation functions with individual training parameters. Although individual parameter allocation increases the total number of training parameters, which may be a concern in small data settings, we exhibit that it improves the adaptability of neural networks and their predictive performance.  
    \item To the best of our knowledge, we are exploring for the first time the effectiveness of adaptive activation functions in applications containing small data sets with fewer than 100 training samples. To conduct a comprehensive investigation, we consider three additive manufacturing problems, with different characteristics such as the number and type of features, as representative of data-austere problems. These problems include the selection of filament materials and 3D printers, as well as the prediction of printability in a complex additive manufacturing problem. We demonstrate that the use of adaptive activation functions is beneficial and justified for these different problems, eliminating the need for predetermined functions. 
    \item Another distinctive feature of our research is the exploration of the effectiveness of adaptive activation functions through the generation of \textit{prediction sets} using conformal inference \cite{shafer2008tutorial,barber2023conformal}, as opposed to relying solely on point predictions. This approach enables us to assess how adaptive activation functions influence the predictive uncertainty of neural network models. In pursuit of this objective, we utilize two metrics: empirical coverage and the average size of the prediction set. In conformal inference, empirical coverage refers to the proportion of actual observed outcomes that fall within the constructed prediction set and the average size gives an indication of the typical range of uncertainty associated with the predictions. A smaller average size suggests more precise and narrow prediction sets, whereas a larger average size indicates a wider range of uncertainty that can be a major consideration in the deployment stage. Hence, this work involves a comprehensive assessment of the reliability of neural networks equipped with adaptive activation functions. This evaluation is particularly crucial in addressing scientific challenges characterized by limited data availability.
    \item We provide source code for the implementation of adaptive activation functions with shared and individual trainable parameters in a hidden layer. The source code, available on GitHub \url{https://github.com/farhad-pourkamali/AdaptiveActivation}, contains a Keras-compatible implementation of three trainable activation functions derived from ELU, Softplus, and Swish. The code provided and our insights will enable practitioners to use adaptive activation functions in a wider range of data-limited problems while eliminating the need to manually set them ahead of time. 
\end{enumerate}

The remainder of this paper is organized as follows. In Section \ref{sec:back}, we discuss some mathematical notations and foundations of neural networks together with popular fixed-shape activation functions. Section \ref{sec:adapt} provides an overview of the existing work on the use of adaptive activation functions. In Section \ref{sec:proposed}, we present the systematic approach we use to rigorously evaluate the predictive performance of adaptive activation functions with shared and individual parameters in small data settings. This evaluation is carried out in three testbeds, each characterized by sparse experimental data originating from additive manufacturing problems.  The last section of this paper provides concluding remarks and outlines potential areas of future study.

\section{Notations and Background Information}\label{sec:back}
An extensively used neural network type for structured or tabular data problems is the fully connected network, commonly called a multilayer perceptron (MLP) \cite{ke2020quality}. This architecture consists of densely interconnected layers organized in a sequential manner, presenting an elegant way to learn nested or composite functions crucial for capturing nonlinear input-output relationships. To be formal, imagine a network with $L$ hidden layers, located between the input and output layers. The $l$-th hidden layer consists of $N_l$ neurons or units. Also, assume that this network takes an input vector $x\in\mathbb{R}^D$, where $D$ is the number of given attributes or features. Furthermore, the weight matrix and the bias vector can be written as $W^{(l)}\in\mathbb{R}^{N_l\times N_{l-1}}$ and $b^{(l)}\in\mathbb{R}^{N_l}$, for each layer indexed by $l=1,\ldots,L+1$. The predicted output $f(x)$ is then defined from the input $x$ according to the following equations:
\begin{align}
    \text{Input layer: }&x^{(0)}=x, \nonumber \\
    \text{Hidden layers: }&x^{(l)}= g^{(l)}\big(\underbrace{W^{(l)}x^{(l-1)}+b^{(l)}}_{z^{(l)}:\text{ weighted sum}}\big),\;l=1,\ldots,L, \nonumber \\
    \text{Output layer: }&f(x) = g^{(L+1)}\big(W^{(L+1)}x^{(L)}+b^{(L+1)}\big).\label{eq:nn}
\end{align}
Therefore, the prediction model $f(x)$ takes on a composite or nested form. The number of units in the output layer $N_{L+1}$ and the corresponding activation function $g^{(L+1)}$ depend on the problem at hand. For example, in binary classification problems, a single neuron is typically placed, i.e., $N_{L+1}=1$, and the Sigmoid activation function is used to find the probability that the data point $x$ belongs to each class. For a given input $z$, the Sigmoid activation function takes the form of: $\text{Sigmoid}(z)=1/(1+e^{-z})$. However, when treating classification problems of more than two classes, the last layer contains one neuron for each class. In this case, we employ the Softmax function to find the categorical distribution for all classes. That is, if we have $C$ classes, then the Softmax function accepts a set of $C$ real-valued numbers $z_1,\ldots,z_C$ and converts them into a valid probability distribution by returning $e^{z_c}/\sum_{c'}e^{z_{c'}}$, for $c=1,\ldots,C$ \cite{ren2020balanced}. 

On the other hand, we have greater flexibility when selecting activation functions for intermediate or hidden layers because they provide ``latent'' representations. Note that in \eqref{eq:nn}, the activation function is applied in the element-wise form, so the standard implementation of dense layers in popular deep learning libraries, such as Keras, follows the same activation function for all neurons in a particular layer. In other words, all neurons placed in a hidden layer apply the same nonlinear function to transfer information to the next layer.  Consequently, significant efforts are required to conduct a comprehensive search to select appropriate levels of nonlinearity in order to maintain input-output relationships.

Initial efforts to train neural networks mainly focused on the hyperbolic tangent (tanh) function to activate hidden layers through $g^{(l)}$, $l=1,\ldots,L$. In the remainder of this paper, we omit the $(l)$ superscript to simplify the notation. The tanh activation function has the following form: 
\begin{equation}
    g(z)=\frac{e^z - e^{-z}}{e^z + e^{-z}}.
\end{equation}
This function has a range of $(-1,1)$, which has an advantage due to the zero-centered structure. However, because this function has two different horizontal asymptotes, the derivatives of this function become very small as we move further away from the origin, which is problematic for gradient optimization techniques. Rectified Linear Unit or ReLU has been widely used in recent years to address the vanishing gradient problem \cite{hu2021handling}. ReLU is defined as $g(z)=\max(z,0)$. Hence, when $z\geq 0$, we have $g(z)=z$ and its derivative is always $1$. On the other hand, we get $g(z)=0$ for negative input values $z$, therefore, the derivative is equal to $0$. Despite the simplicity of ReLU, it exhibits a saturating region, which can be problematic for gradient descent optimization. Specifically, ReLU discards negative values, leading to the known problem of dying ReLU \cite{yang2023dprelu}. Consequently, several activation functions have been devised to address these concerns while maintaining the fundamental structure of ReLU. 

Exponential Linear Unit (ELU) shares a structure similar to ReLU for nonnegative inputs $z$, yet it facilitates the flow of information to some degree for negative values of $z$ through the expression $(e^z-1)$. Additionally, Softplus is another popular activation function that provides a smooth approximation of ReLU in the form of $g(z)=\log(e^z+1)$ for all values of $z$. Consequently, for sufficiently large positive $z$ values, this function mimics a linear behavior, i.e., $g(z) \approx z$, owing to the inverse relationship between the logarithmic and exponential functions. Another way of approximating the ReLU activation function was proposed in \cite{ramachandran2017searching}, named the Swish activation function, which takes the form of $g(z)=z\cdot\text{Sigmoid}(z)=z/(1+e^{-z})$. Similar to ReLU, Swish acts as a linear function for large positive $z$ values because the denominator becomes very close to $1$. However, what sets Swish apart is its departure from the widely-used monotonicity property. Building on the empirical success of Swish in various computer vision problems, several Swish variants have been proposed in recent years, as highlighted in \cite{zhu2021logish}.

\section{Existing Work on Adaptive Activation Functions}\label{sec:adapt}

In recent advancements in neural networks, there has been a growing interest in exploring trainable or adaptive activation functions to enhance their flexibility and adaptability. Traditional activation functions that we discussed in the previous section have fixed shapes, and their performance may be suboptimal for certain tasks. Thus, users must make significant efforts to choose an appropriate activation function from the collection of existing built-in functions. By introducing adaptive activation functions, neural networks can progressively learn the optimal form of input-output relationship for each unit or layer and accelerate the training process. To elucidate this concept, consider the standard empirical risk minimization problem for model fitting with fixed activation functions:
\begin{equation}
\theta^*\in\argmin_{\theta} \sum_{n=1}^{N_{\text{train}}} \text{loss}\big(y_n, f_{\theta}(x_n)\big),\;\theta:=\big\{W^{(1)}, b^{(1)},\ldots,W^{(L+1)}, b^{(L+1)}\big\},\label{eq:ERM}
\end{equation}
where $\mathcal{D}_{\text{train}}=\{(x_n,y_n)\}_{n=1}^{N_{\text{train}}}$ represents the training data set comprising feature and response pairs. The loss function serves as a metric to evaluate the accuracy of predictions, employing specific formulations like the quadratic loss function in regression problems and cross-entropy in classification tasks. In the optimization process, the variable $\theta$ serves as a container for weight matrices $W^{(l)}$ and bias vectors $b^{(l)}$, $l=1,\ldots,L+1$. These values are obtained through an initialization step, followed by an iterative refinement of their values. The refinement entails computing the gradient of the loss function with respect to the elements of $\theta$ and applying gradient descent to minimize the loss function over iterations.

The core concept of adaptive activation functions lies in \textit{parameterizing} activation functions to learn the optimal form of nonlinearity introduced in \eqref{eq:nn}.  To this end, let $g(z;\alpha)$ be a parameterized activation function and assume that we utilize $N_{h}$ distinct activation functions in a neural network model. In this case, we can modify the optimization variable in \eqref{eq:ERM} by adding the new parameters, i.e., $\theta=\{W^{(1)}, b^{(1)},\ldots,W^{(L+1)}, b^{(L+1)}, \alpha_1,\ldots,\alpha_{N_h}\}$. As a result, the total number of parameters to be learned through gradient descent increases, prompting the key question of whether the set of enhanced parameters can improve predictive power compared to fixed-shape activation functions characterized by predefined values of $\alpha$. 

In this work, we implement and examine some popular parameterized activation functions, starting with the trainable Exponential Linear Unit (ELU) function, which has the following form:
\begin{equation}
\text{ELU}(z;\alpha)=\begin{cases}z & \text{if } z\geq 0 \\ \alpha(e^z - 1) & \text{if } z< 0\end{cases}.
\end{equation}
In many deep learning libraries, the default value for $\alpha$ is set to $1$. The derivative of the parameterized ELU is given by $\alpha e^z$ for negative values of $z$, making the selection of $\alpha$ a crucial consideration during the training process. For instance, when dealing with negative values of $z$, where $e^z$ tends to be very small, opting for a larger value of $\alpha$ becomes essential to prevent the derivative from approaching zero.

Similarly, the parameterized version of Softplus can be written as $\text{Softplus}(z;\alpha)=\log(e^z + \alpha^2)$. While the default value of $1$ results in a smooth approximation of ReLU, choosing very small values for $\alpha$ renders this activation function nearly linear. This attribute plays a crucial role in finely regulating the complexity of the mapping accomplished by every hidden layer. Lastly, we can introduce an additional tuning parameter for the Swish activation function in the form of $\text{Swish}(z;\alpha)=z\cdot \text{Sigmoid}(\alpha z)=z/(1+e^{-\alpha z})$. Notably, when $\alpha=0$, this activation function behaves as a scaled linear function, and gradually converging to ReLU as $\alpha$ increases. Furthermore, we can show that the derivative of the parameterized Swish activation function has the following form:
\begin{equation}
\frac{d}{dz}\text{Swish}(z;\alpha)= (1+\alpha z)\cdot\text{Sigmoid}(\alpha z) - \alpha z \cdot \big(\text{Sigmoid}(\alpha z)\big)^2. 
\end{equation}
Thus, setting $\alpha$ to 0 results in a constant derivative of 0.5. Conversely, when $\alpha$ takes nonzero values, the derivative of Swish around the origin becomes significantly dependent on the chosen $\alpha$.

\begin{figure}[ht]
    \centering
    \includegraphics[width=\textwidth]{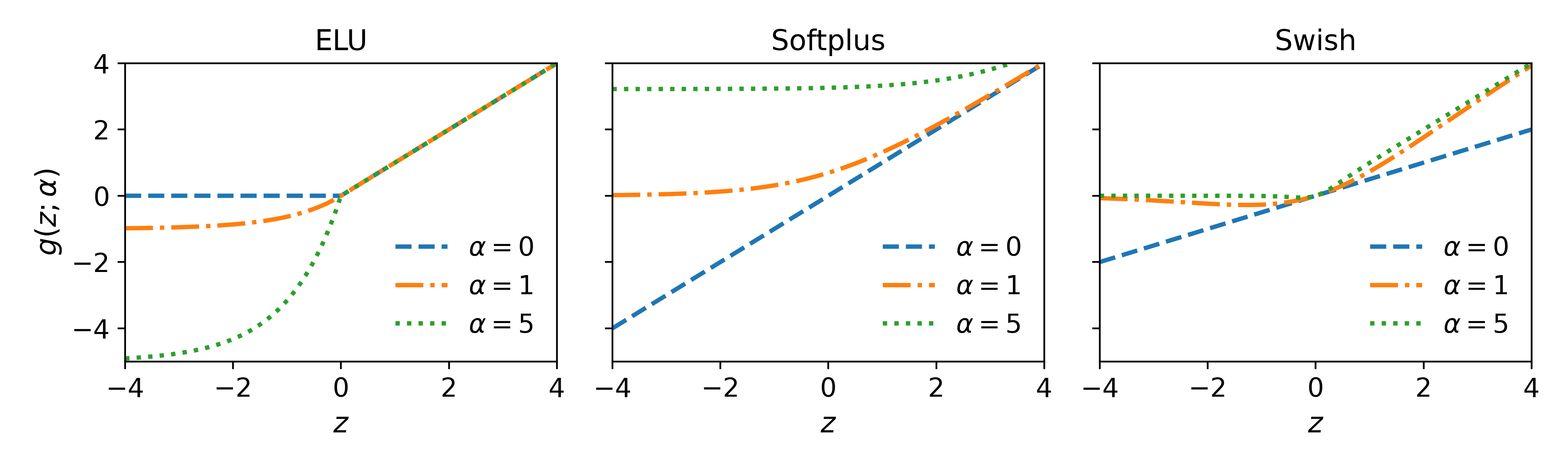}
    \caption{Visualizing the impact of the ``trainable'' parameter $\alpha$ on modifying the structure of three widely-used activation functions: Exponential Linear Unit (ELU), Softplus, and Swish. The default value for $\alpha$ in fixed activation functions is commonly set to $1$.}
    \label{fig:activation}
\end{figure}

Figure \ref{fig:activation} depicts the three parameterized activation functions, showcasing their distinctions across a range of $\alpha$ values to provide a visual understanding of their differences. Setting $\alpha=0$ in ELU yields the well-known ReLU activation function. However, by increasing $\alpha$, ELU enables the information flow for negative values of $z$. It should be noted that for nonnegative input values $z$, the parameter $\alpha$ does not alter the structure of ELU. In contrast, $\alpha$ plays a pivotal role in shaping the overall characteristics of Softplus and Swish for all values of $z$.

Recent research dedicated to investigating the effectiveness of adaptive activation functions has predominantly focused on the analysis of image data sets within the domain of computer vision \cite{ccatalbacs2023deep}. A comprehensive survey by Apicella et al. \cite{apicella2021survey} synthesized findings from around 20 studies, extensively utilizing established benchmarks such as MNIST, CIFAR10, CIFAR100, and ImageNet. For example, on the CIFAR10 data set, the median classification accuracy scores using the fixed ReLU and ELU activation functions were approximately $0.91$ and $0.93$, respectively. However, employing adaptive ELU and Swish resulted in slightly higher accuracy scores of $0.94$ and $0.95$, respectively. Additionally, recent papers by Dubey et al. \cite{dubey2022activation} and Emanuel et al. \cite{emanuel2023effect} delved into the exploration of adaptive activation functions, specifically for tasks related to language translation, speech recognition, and text classification.

Expanding the domain, Wang et al. \cite{wang2022kdac} extended the scope by exploring the application of adaptive activation functions across six benchmark data sets in the social and e-commerce domains. Moreover, Klopries et al. \cite{klopries2023flexible} investigated adaptive activation functions in the context of unsupervised learning for training autoencoders. In addition, Jagtap et al. \cite{jagtap2023important} and another recent work \cite{10227556} contributed to the literature by studying adaptive activation functions in the context of solving various forward and inverse problems within the framework of physics-informed neural networks, where the goal is to combine the expressive power of neural networks with the physical constraints provided by partial differential equations. 

Despite the abundance of research in the area of adaptive activation functions, there is a notable gap in understanding the effectiveness of adaptive activation functions in addressing sparse data challenges, particularly those arising from intricate experimental scenarios within scientific and engineering domains. To bridge this gap, this paper concentrates on three distinct experimental data sets derived from various additive manufacturing problems. Improving the predictive performance of neural networks and reducing the need for predetermined activation functions are indispensable to accelerate the design and discovery of novel materials with enhanced flexibility and efficiency. 

In our study, we specifically evaluate the efficacy of adaptive activation functions in scenarios with fewer than $100$ training samples. This focus provides a unique perspective compared to existing research, which has predominantly focused on data sets with significantly larger sizes spanning multiple orders of magnitude. To illustrate this point, consider the CIFAR10 data set\textemdash a collection of 60,000 32x32 color images distributed across 10 classes, with 6,000 images per class. In contrast, in many experimental setups, the average sample size per class is on the order of a few tens of samples. It is worth noting that the expense of gathering a few tens of samples in the world of material experiments may significantly surpass the cost of collecting and labeling images with sizes several orders of magnitudes larger due to machine, maintenance, material, and labor costs. As a result, there is a compelling need for new research to understand the trade-off between performance gains and the increase in trainable parameters associated with the utilization of adaptive activation functions.

\section{Adaptive Activation Functions in Small Data Settings}\label{sec:proposed}
In this section, we undertake a comprehensive evaluation of adaptive activation functions using three distinct experimental data sets. Additionally, we explore three activation functions\textemdash ELU, Softplus, and Swish\textemdash utilizing both fixed and trainable parameters, denoted as $\alpha$. As mentioned earlier, fixed activation functions correspond to the specific choice of $\alpha=1$, while adaptive activation functions dynamically learn and optimize the optimal value of $\alpha$ through gradient descent optimization during the training process. To ensure a fair comparison, we initialize the optimization process with a value of $1$ for $\alpha$ and then report the optimized value of $\alpha$ after completing the training stage. Furthermore, we configured the number of epochs to be 100, employing the Adam optimization algorithm with a learning rate of $0.05$.

In terms of the neural network architecture, our focus in this paper is on MLPs consisting of a single hidden layer, a deliberate choice driven by the constraint of a small sample size. This choice of $L=1$ also enables us to explore a critical aspect of adaptive activation functions that has been overlooked in previous research. To elucidate, we examine the common scenario of identical activation functions, where all units in a layer share the same $\alpha$ parameter. However, we go a step further and investigate the impact of allocating individual parameters for units within a hidden layer. This exploration aims to understand the trade-off between the number of trainable parameters and the adaptability of neural networks, particularly in small data settings. 

\begin{figure}[ht]
    \centering
    \includegraphics[width=\textwidth]{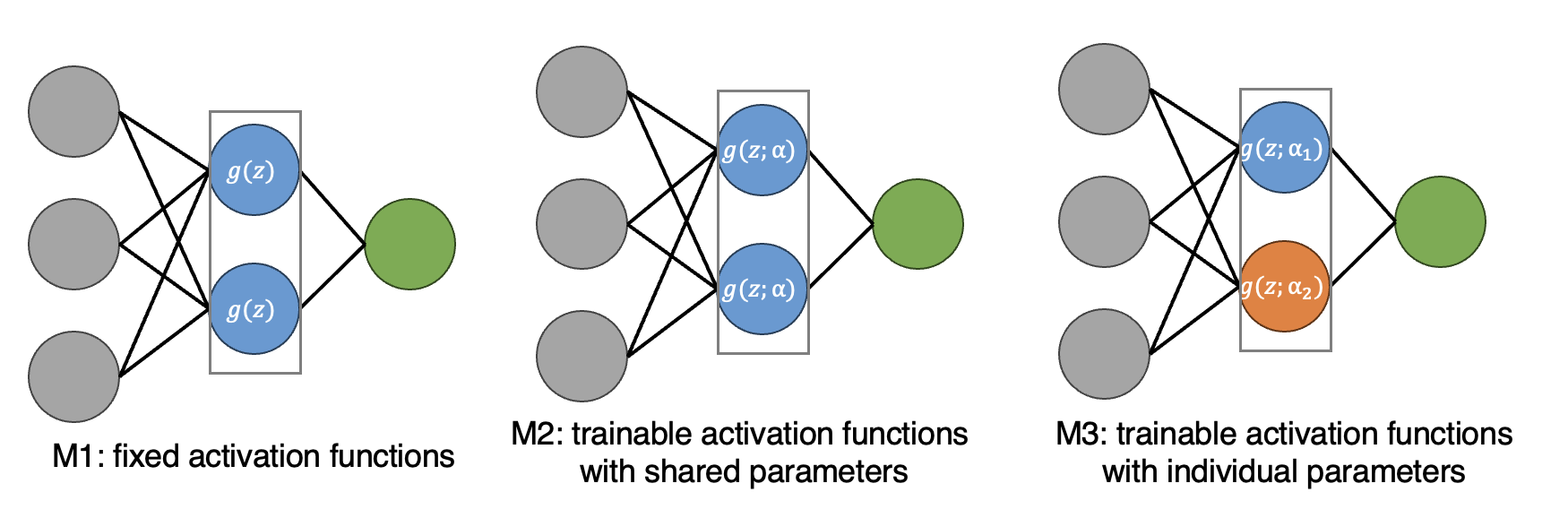}
    \caption{Demonstrating the spectrum of flexibility within the neural network models under examination. M1 denotes the conventional fixed activation functions, while M2 permits a single trainable parameter for the hidden layer. Conversely, M3 offers the utmost flexibility by assigning an individual trainable parameter to each unit in the hidden layer. To implement M3, we employ the Keras Functional API, connecting each hidden unit to the input layer and subsequently concatenating their outputs to form a unified hidden layer. While we fix the number of hidden units $N_h=2$, the number of units in the input and output layers are determined by the characteristics of the labeled data specific to each additive manufacturing problem that we consider in this paper.}
    \label{fig:architecture}
\end{figure}

Figure \ref{fig:architecture} depicts our systematic investigation, where the model M1 represents networks with fixed activation functions, M2 corresponds to networks that share the same trainable parameter $\alpha$, and M3 allows for individual trainable parameters for each activation function. In our investigation, we set the number of units in the hidden layer to be $N_h=2$ due to the limited availability of data. However, for completeness, we will explore the impact of $N_h$ on the performance gap between neural networks with adaptive and fixed activation functions at the end of this section.

To gauge the influence of employing trainable activation functions on the confidence of prediction models, we propose a shift from the conventional evaluation method, which relies on singular point predictions, to the creation and assessment of prediction sets. Formally, consider a classification problem with $C$ classes, denoting the output space as $\mathcal{Y}={1,\ldots,C}$. Adopting the conformal inference framework \cite{shafer2008tutorial,angelopoulos2023conformal}, our objective is to construct a prediction set for a new or unseen input vector $x_{\text{test}}$, expressed as $\mathcal{I}(x_{\text{test}})\subseteq \mathcal{Y}$, adhering to the following probabilistic rationale:
\begin{equation}
    \text{Prob}(y_{\text{test}}\in\mathcal{I}(x_{\text{test}})\geq 1-\delta,\;\delta\in(0,1)
\end{equation}
where the desired coverage level $1-\delta$ is set by users. In this paper, we opt for the conventional selection of $\delta=0.1$ to establish the desired coverage level at a reasonable level $0.9$. 

Next, we discuss an efficient algorithm to find the prediction set in our numerical experiments. During the training stage, we define the score function as one minus the softmax score of the true class. That is, we have $s(x_n,y_n)=1-f(x_n)_{y_n}$, where $f(x_n)_{y_n}$ represents the softmax score of the true class indicated by the ground-truth label $y_n$. Then, we find the empirical quantile of $\{s(x_n,y_n)\}_{n=1}^{N_{\text{train}}}$ at the adjusted level $\lceil (1-\delta)(N_{\text{train}} +1)/ N_{\text{train}}\rceil$ \cite{angelopoulos2023conformal}. Once we obtain the quantile $\hat{q}$, the prediction set for the test vector $x_{\text{test}}$ can be found as follows: 
\begin{equation}
\mathcal{I}(x_{\text{test}})=\big\{y: f(x_{\text{test}})_y\geq 1-\hat{q}\big\}.
\end{equation}
This implies that the prediction set encompasses all classes with softmax scores that exceed the threshold of $1-\hat{q}$.

Given these prediction sets, we evaluate the impact of adaptive activation functions using two metrics: empirical coverage and uncertainty score. Let $\mathcal{D}_{\text{test}}$ represent a distinct testing set with $N_{\text{test}}$ input-output pairs for evaluation. \textit{Empirical coverage} denotes the proportion of data points with true outputs within the corresponding prediction sets. Similar to classification accuracy, higher values are preferable, and we aim for a minimum coverage level of $0.9$. On the other hand, the \textit{uncertainty score} in this context is the average size of prediction sets. A perfect score is $1$ for confident and accurate classifiers that predict a single class while adhering to the probabilistic argument. However, higher scores are less desirable, indicating a classifier with less confidence in its predictions.

In the following sections, we partition $30\%$ of the data at random to construct the testing set $\mathcal{D}_{\text{test}}$, while the remaining portion is allocated for forming the training data set $\mathcal{D}_{\text{train}}$. Recognizing the sensitivity of neural networks to data splits, we conduct $20$ independent splits and visualize the distribution of evaluation metrics across these experiments. Our chosen visualization method is the violin plot, which combines key elements from box plots and kernel density plots. The width of this plot signifies the density of data points at various values, and the vertical axis represents the probability density. Additionally, the tick in a violin plot denotes the median of each evaluation metric.

As a final note before delving into our three testbeds, it is important to highlight that the version of conformal prediction discussed here differs slightly from the split conformal prediction methods. The key distinction lies in the fact that split conformal prediction typically employs a separate calibration data set for quantile determination. However, in this paper, we opt not to divide the training set for two reasons. First, across all testbeds, our training samples are fewer than 100, and the objective is to utilize as many data points as possible for training neural networks, including optimizing trainable parameters for each activation function. Second, utilizing the training data set enables us to indirectly assess the risk of overfitting\textemdash a significant concern in predictive modeling with neural networks. The discussed approach allows us to account for artificially small score functions due to overfitting, negatively impacting empirical coverage on the test set, a factor considered in our comprehensive analysis.

\subsection{Filament Selection}
Fused filament fabrication, a form of material extrusion additive manufacturing, has gained popularity due to its cost effectiveness, ease of operation, and compatibility with a wide range of thermoplastics \cite{lee2019fabrication,WU2018298,pei2022combining}. One of the challenges associated with fused filament fabrication is that the quality of the final product depends on many design and processing parameters, including but not limited to layer height, extrusion temperature, print bed temperature, infill density, and infill pattern \cite{goh2020process}.  Therefore, the selection of appropriate materials and processing and design conditions for a given application is a significant challenge. For example, trial-and-error can be used to determine the upper and lower bounds of adjustment for a parameter. Once these bounds are known, a linear iterative optimization approach can be used to navigate between them. However, machine learning-based surrogates offer the potential to significantly accelerate this search process.

To demonstrate the effectiveness of our classification models in addressing a comparable issue of selecting an appropriate material for fused filament fabrication, we opted for a data set comprising 11 input features. These features encompass four design parameters: layer height, wall thickness, infill density, and infill pattern. Four process parameters (extrusion temperature, print bed temperature, print speed, and fan speed) were also considered. Lastly, three material parameters (roughness, tensile strength, and elongation at break) were included in the data set. The objective of the models is to accurately characterize the material as polylactic Acid (PLA) or acrylonitrile butadiene styrene (ABS).
This data set contains $70$ labeled data points, which can be accessed at \cite{apmonitor}. 

\begin{figure}[ht]
    \centering
    \includegraphics[width=\textwidth]{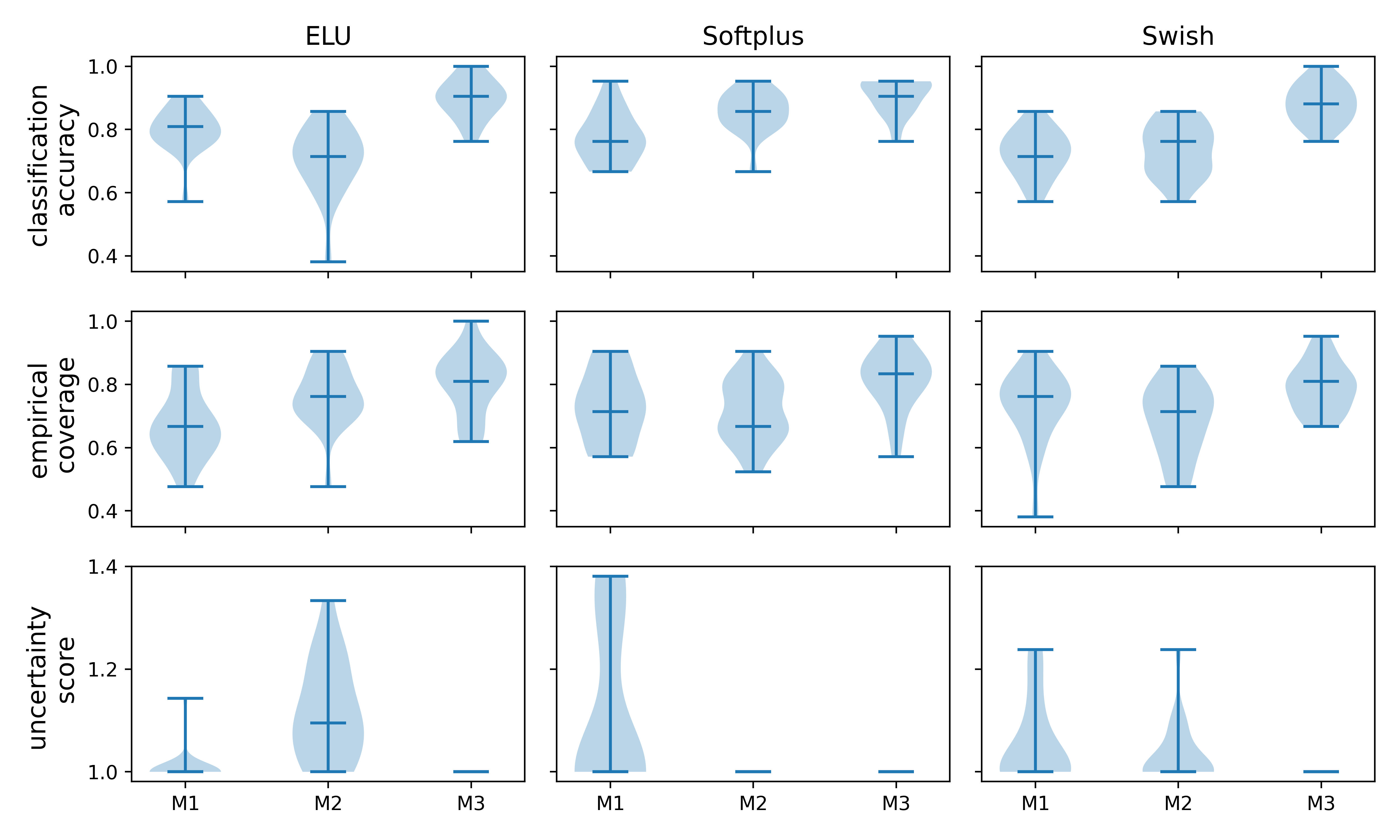}
    \caption{Employing the filament selection problem as a benchmark, we assess the performance of M1, M2, and M3 using three evaluation metrics. Classification accuracy denotes the fraction of correct predictions on the test data set, while empirical coverage and uncertainty score are derived from prediction sets within the conformal prediction framework using $\delta=0.1$. We note that M3 demonstrates superior performance compared to M1 and M2. Notably,  the worst-case classification accuracy score produced by M3 is comparable to the median score attained by both M1 and M2.}
    \label{fig:filament}
\end{figure}

In Figure \ref{fig:filament}, we assess the performance of the three discussed models\textemdash M1, M2, and M3\textemdash utilizing three evaluation metrics: classification accuracy, empirical coverage, and uncertainty score (i.e., the average size of the prediction set). Based on the first row of Figure \ref{fig:filament} that reports the classification accuracy results using plain point predictions, we see that the median scores for M1 with fixed ELU, Softplus, and Swish activation functions are $0.80$, $0.76$, and $0.71$, respectively. Although the best median score is achieved using fixed ELU when M1 is under investigation, a significant drawback of ELU is that its worst-case performance or the minimum score in $20$ data splits is $0.57$, which is comparable to that of the Swish activation function. Furthermore, the second and third rows of this figure reveal that empirical coverage obtained using the test set $\mathcal{D}_{\text{test}}$ is consistently less than the target coverage level of $0.9$. For example, the median value of empirical coverage for fixed ELU is $0.66$. It is also observed that M1 with the fixed Swish activation function exhibits suboptimal empirical coverage, reaching as low as $0.38$. Therefore, in this case study, we conclude that the three fixed activation functions do not provide accurate and reliable classification models to predict the filament type. 

Surprisingly, we observe that the conventional practice of sharing the same trainable parameter across all units in a hidden layer, as implemented in M2, does not produce noticeable improvements compared to M1 with fixed activation functions. For example, the median classification accuracy scores achieved by M2 using trainable ELU, Softplus, and Swish are $0.71$, $0.85$, and $0.76$, respectively. This suggests that the most significant positive impact is observed for adaptive Softplus, while the introduction of the trainable parameter $\alpha$ decreases the overall accuracy score of the ELU activation function. We notice similar patterns for the second and third rows of Figure \ref{fig:filament} when M2 is under investigation. Although the empirical coverage for adaptive ELU is marginally better than that for fixed ELU, it is accompanied by a higher uncertainty score. This indicates that the classifier employing adaptive ELU experiences elevated levels of uncertainty compared to its fixed counterpart. In addition, M2 does not result in notable enhancements in empirical coverage or reductions in the uncertainty score for Softplus and Swish.

The next step involves examining the performance of M3, which incorporates adaptive activation functions with individual trainable parameters. This model shows notable improvements compared to both M1 and M2 from various perspectives. To begin with, as illustrated in the first row of Figure \ref{fig:filament}, the median classification accuracy scores hover around $0.9$ for the three selected activation functions. Also, it is important to note that the worst-case performance of M3 is $0.76$, which is comparable to the median scores obtained by M1 and M2.  On the other hand, M3 achieves the best classification accuracy scores of $1$, $0.95$, and $1$ when employing ELU, Softplus, and Swish, respectively. Consequently, the integration of adaptive activation functions with individual parameters in M3 represents a substantial improvement in terms of the number of correct predictions. Therefore, our investigation indicates that introducing additional parameters in the training stage is beneficial, even with the limited availability of labeled data.  

Shifting focus to conformal prediction for assessing predictive uncertainty beyond plain point predictions, M3 consistently provides superior empirical coverage compared to M1 and M2. For example, the median values of empirical coverage obtained by M3 for ELU, Softplus, and Swish are $0.80$, $0.83$, and $0.80$, respectively. Although the lower-than-expected values of empirical coverage can be attributed to utilizing training samples as calibration data points, it is noteworthy that M3 consistently delivers the highest-quality prediction sets. Importantly, we observe that the constructed prediction sets contain only a single class, indicating that the use of adaptive activation functions with individual trainable parameters results in accurate and reliable classifiers in this example.

\subsection{Printer Selection}
In addition to the choice of design, material, and printing parameters, the final properties of the parts produced by fused filament fabrication also depend on the specific 3D printer used \cite{braconnier2020processing}. This dependence is due to variations in hardware and firmware among different printer models, which can affect material deposition rate, movement precision, and temperature control. Understanding these printer-to-printer variations requires insight into the printing process. Fused filament fabrication begins with slicing three-dimensional computer-aided designs into two-dimensional layers. The printer nozzle then deposits material onto the build surface sequentially, layer-by-layer. 

Despite using similar printing parameters (e.g., nozzle temperature, print bed temperature, layer height, infill pattern, and infill density) and the same print geometry, different printers may follow distinct toolpaths, as dictated by both the slicing software and the printer's own embedded systems. The resulting differences in layer time and deposition pattern significantly impact interlayer adhesion, a key factor for the mechanical properties of the printed object, leading to variations in the strength and surface properties of the final product \cite{gao2021fused}. Consequently, the process-structure-property correlations also vary across different printers.

In this section, we assess the performance of our classification models, which are designed to identify the specific 3D printers used to manufacture parts. This evaluation is based on a data set collected by Braconnier et al. \cite{braconnier2020processing} that contains a total of $104$ input-output pairs. This data set comprises tensile properties\textemdash specifically, tensile strength, elastic modulus, and elongation at break\textemdash of parts fabricated using three different 3D printers: MakerBot Replicator 2X, Ultimaker 3, and Zortrax M200. These parts were produced with variations in extrusion temperature, layer height, print bed temperature, and print speed. Therefore, our classification models aim to predict the type of printer used for each print, considering $7$ input features: tensile strength, elastic modulus, elongation at break, extrusion temperature, layer height, print bed temperature, and print speed.

\begin{figure}[ht]
    \centering
    \includegraphics[width=\textwidth]{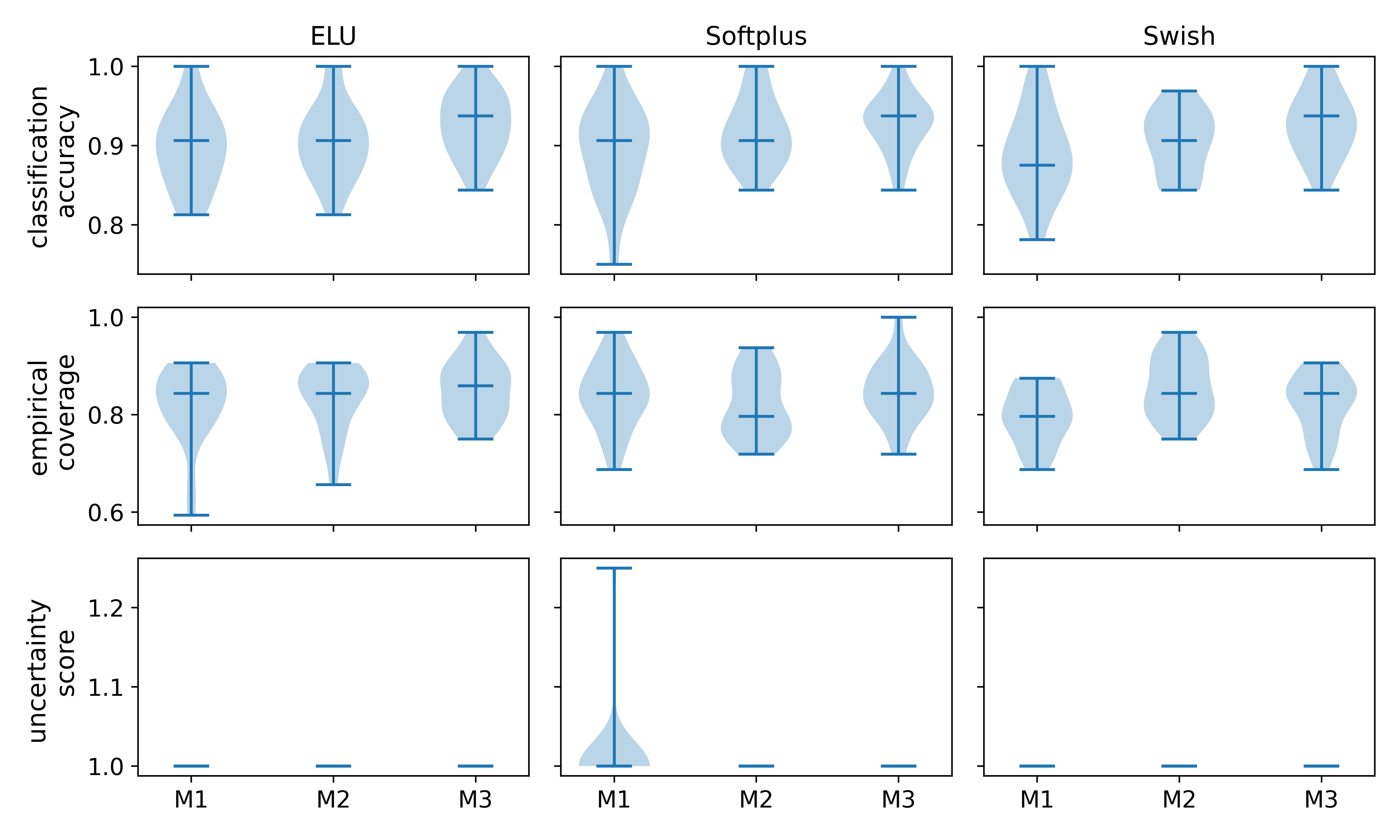}
    \caption{Employing the printer selection problem as a benchmark, we assess the performance of M1, M2, and M3 using three evaluation metrics. Using the trainable ELU activation function with individual parameters in M3 yields the highest classification accuracy and empirical coverage scores. According to the information from the third row, all prediction sets consist of a single class, except for the fixed Softplus activation function in M1.}
    \label{fig:printer}
\end{figure}

In Figure \ref{fig:printer}, we show the comparison results for M1, M2, and M3 with varying levels of adaptability using the printer selection data set. According to the first row of this figure, which presents the conventional classification accuracy score, both M1 and M2 exhibit median scores of approximately $0.9$. Notably, in this comparison, when utilizing Softplus and Swish, the worst-case scores for M2 are noticeably superior to those of M1. For example, the lowest accuracy score for Softplus with a shared trainable parameter is $0.84$, while the corresponding value of the fixed Softplus activation function is $0.75$. Furthermore, we find consistent results within the conformal prediction framework. In general, M2 provides better empirical coverage compared to M1. For example, the maximum empirical coverage values obtained by M2 using ELU, Softplus, and Swish are $0.90$, $0.93$, and $0.96$, respectively. At the same time, the prediction sets generated using M2 contain only a single class, further strengthening the confidence in the prediction models. Therefore, in this case study, adaptive activation functions with a shared trainable parameter can achieve the target coverage level of $0.9$. 

Furthermore, Figure \ref{fig:printer} indicates that the utilization of individual trainable parameters in M3 leads to higher classification accuracy scores when compared to M2. In particular, the median accuracy score for all three activation function choices in M3 is about $0.93$. Additionally, the worst-case accuracy score in M3 reaches $0.84$, which is comparable or superior to that of M2. Consequently, our investigation demonstrates that the increased flexibility of activation functions in M3 improves the classification accuracy score, even with a small sample size. 

Within the conformal prediction framework, we also observe substantial improvements in empirical coverage when using individual trainable parameters along with ELU and Softplus. For example, the maximum empirical coverage scores for ELU and Softplus are $0.96$ and $1$, respectively. Therefore, the prediction models in M3 can achieve the desired coverage level of $0.9$. As a result, this experiment demonstrates that the use of fully adaptive activation functions in M3 provides more accurate and reliable prediction models to determine the type of 3D printer compared to M1 and M2.

Lastly, this experiment highlights the importance of employing the conformal prediction framework for a thorough evaluation of classification models. Although the conventional point predictions indicate that the use of Swish in M3 yields high classification accuracy, a closer look at the second row of Figure \ref{fig:printer} reveals that the maximum and minimum empirical coverage scores for Swish in M3 fall short compared to ELU and Softplus. For example, the minimum empirical coverage score in M3 using Swish is $0.68$, while ELU and Softplus enjoy values of $0.75$ and $0.71$ that are closer to the target coverage level. In light of our uncertainty-aware evaluation approach, it becomes evident that ELU delivers the best overall performance.

\subsection{Printability Prediction}
Vat photopolymerization additive manufacturing fabricates parts by selectively curing a photopolymer feedstock using a light source, typically ultraviolet light. One exciting use of vat photopolymerization is in creation of highly filled polymer composites \cite{shah2021highly,zakeri2020comprehensive,wang2020fabrication}. Increasing the solid filler content in the feedstock suspension can enhance properties of printed parts, including compressive and tensile strength \cite{al2021vat}. However, increasing the solid content beyond 35 vol.\% makes the printing process challenging due to increases in viscosity and light scattering. Increasing the filler amount in the suspension causes a exponential increase in viscosity, which can hinder the recoating process between layers, resulting in weak interlayer adhesion, defective prints, or even print failure \cite{konijn2014experimental}.

To address the viscosity issue corresponding to highly filled (50 vol.\% to 70 vol.\%) suspensions, a bimodal system can be utilized \cite{delarue2023increasing}. This approach involves mixing two distinct sizes of solid particles, each with a different particle size distribution, to enhance packing density.  However, experimentally optimizing the blend ratios of fine to coarse particles for minimal viscosity is a resource-intensive process. Additionally, increased solid content can reduce the cure depth, the extent to which the light effectively cures the suspension, due to light scattering \cite{tomeckova2010critical,tomeckova2010cure}. This scatter will lead to prolonged printing times. Therefore, the energy to cure the print layers often needs to be increased. However, excessive curing energy introduces the risks overcuring and parts adhering to the vat, causing print failures. Conversely, insufficient curing energy may lead to incomplete prints. Hence, cure energies must be optimized for obtaining successful prints. Nevertheless, optimizing the cure energy for highly filled suspensions typically relies on a trial-and-error approach.

In this section, we demonstrate the efficacy of our predictive models in predicting the printability of highly filled (50 vol.\% to 70 vol.\%) bimodal suspensions using a digital light processing-based vat photopolymerization technique. The initial data set includes two input features related to the suspension formulation: the solid loading and blend ratio. Additionally, the data set includes the first layer cure energy, which is the energy used to cure the first five layers onto the build head, and the model layer cure energy, which is the energy input for curing the subsequent layers. The printability of these formulations is labeled as either a ``fail'' or ``success'' for $63$ data points. 

\begin{figure}[ht]
    \centering
    \includegraphics[width=\textwidth]{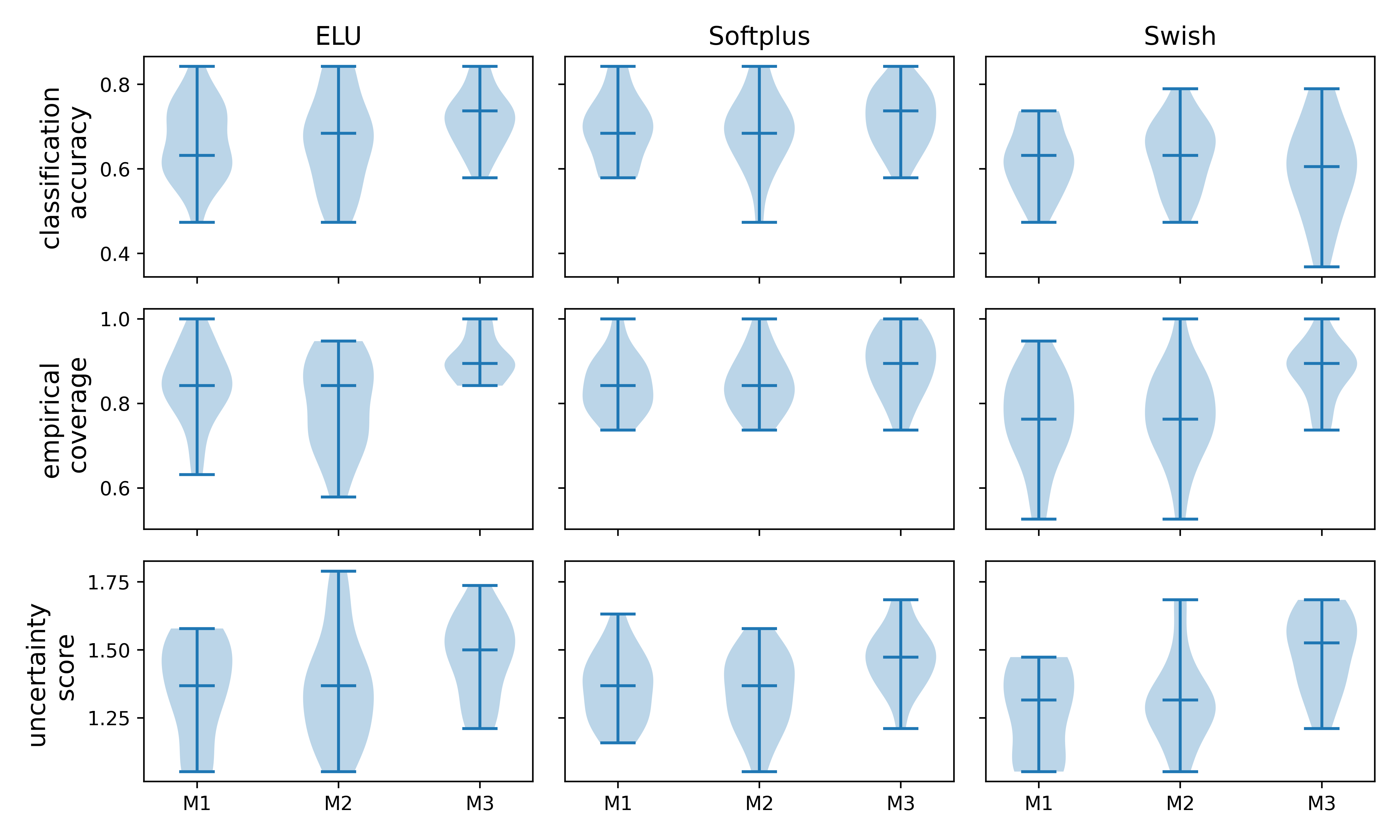}
    \caption{Employing the printability prediction problem as a benchmark, we assess the performance of M1, M2, and M3 using three evaluation metrics. Using the trainable ELU and Softplus activation functions with individual parameters in M3 yields the highest classification accuracy and empirical coverage scores. However, it is worth noting that the minimum classification accuracy score for Swish in M3 is $0.37$, which does not meet the threshold for a random classifier.}
    \label{fig:loading}
\end{figure}

The left column of Figure \ref{fig:loading}, which compares the fixed and trainable version of ELU, demonstrates that the default value of $\alpha=1$ in M1 is inappropriate because the minimum classification accuracy score is about $0.47$, which is below $0.5$, a value that can be achieved by a random classifier. Opting for an adaptive ELU with shared trainable parameters in M2 shows a slight improvement in overall performance since the median classification accuracy score is approximately $0.68$. However, the worst-case performance of ELU in M2 aligns more closely with that of a random classifier.

Interestingly, we note that the use of fully adaptive ELU activation functions in M3 provides further enhancements. For example, the maximum, median, and minimum classification accuracy scores are $0.84$, $0.74$, and $0.58$, respectively. Therefore, the worst-case performance of M3 across 20 data splits is considerably more reasonable. Additionally, the median empirical coverage score of adaptive ELU in M3 stands at $0.89$, closely aligning with the desired coverage level of $0.9$. These data suggest that the predictions made by ELU in M3 are more accurate and reliable compared to M1 and M2.

Furthermore, the middle column of Figure \ref{fig:loading} illustrates that the fully adaptive Softplus activation function in M3 surpasses the performance of M1 and M2. In particular, the median classification accuracy and empirical coverage values for Softplus in M3 are $0.73$ and $0.89$, respectively. Consequently, similar to ELU, Softplus produces accurate and reliable prediction models when each hidden unit is given the flexibility to optimize the structure of its activation function. However, the right column of this figure highlights a significant drawback of fully adaptive Swish in M3, as its minimum classification accuracy score falls below the threshold of $0.5$ that a random classifier can achieve.  On the other hand, the maximum classification scores for Swish in M1 and M2 are below the scores obtained by ELU and Softplus in M3. Therefore, we can conclude that the fully adaptive ELU and Softplus in M3 offer the best overall performance.

\subsection{Exploring the Impact of Optimized $\alpha$ and $N_h$}
\begin{figure}[ht]
    \centering
    \includegraphics[width=\textwidth]{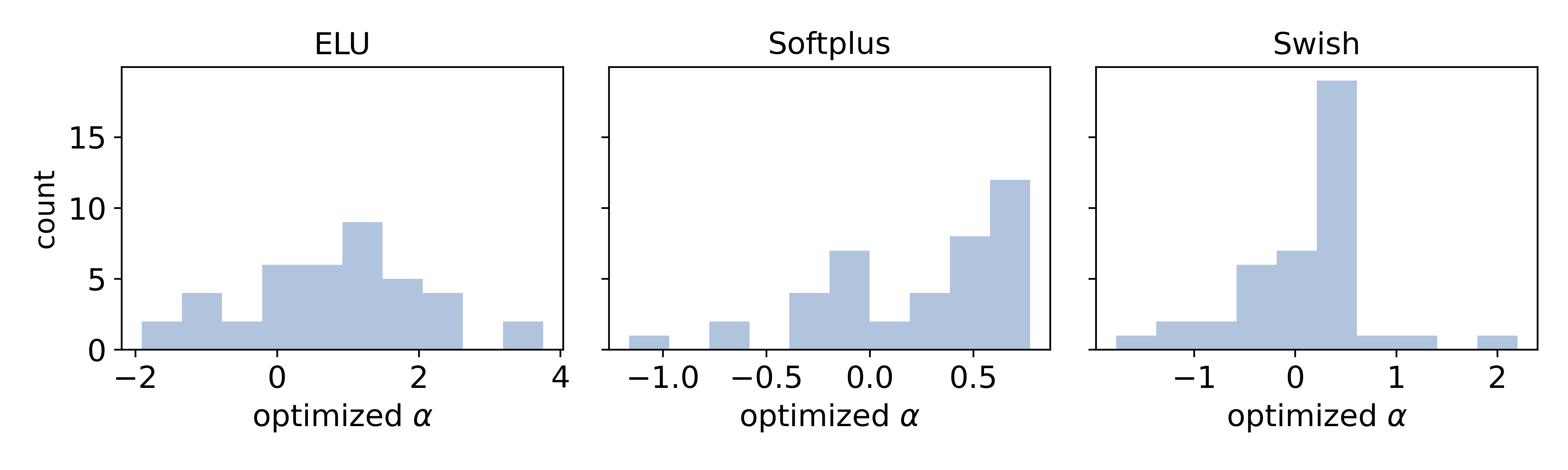}
    \caption{Reporting the learned values of the parameter $\alpha$ used to regulate the structure of each activation function in M3, as derived from the filament selection data set.}
    \label{fig:filament_alpha}
\end{figure}

In this section, our objective is to delve into a more comprehensive understanding of the inner mechanisms governing adaptive activation functions, utilizing the filament selection testbed. The rationale for choosing this particular data set lies in its inclusion of $11$ features, representing the highest count among the three additive manufacturing problems we have examined. In Figure \ref{fig:filament_alpha}, we present histogram plots that show the optimized or learned values of $\alpha$ in M3. It is essential to recall that the parameter $\alpha$ is initially set to $1$, ensuring that the initial structure of all activation functions is aligned with their fixed counterparts. However, we iteratively update the $\alpha$ parameter of each activation function during the training process, in conjunction with other weight matrices and bias terms. Also, note that in this scenario, we have $2$ individual trainable parameters per run. We consider $20$ independent random data splits, resulting in a total of $40$ optimized values.

The histogram plot for adaptive ELU reveals that the optimal value of $\alpha$ is approximately centered around 1, mirroring the default value for its fixed counterpart. However, given the substantial increase in the classification accuracy score in Figure \ref{fig:filament} when using the trainable ELU activation function compared to the fixed ELU, we observe instances where adjusting the information flow for negative inputs $z$ becomes beneficial. For example, the optimized value of $\alpha$ can reach as high as $4$. Notably, we find that the optimized value of $\alpha$ may even assume negative values. This aligns with the observation that the introduction of nonmonotonicity into the activation function can enhance overall predictive performance, facilitating a more nuanced capture of input-output relationships.

The middle graph in Figure \ref{fig:filament_alpha} for the parameterized Softplus activation function, i.e., $\text{Softplus}(z;\alpha)=\log(e^z+\alpha^2)$, demonstrates that the optimized values of $\alpha^2$ are typically in the interval between $0$ and $1$. As depicted in Figure \ref{fig:activation}, it is noteworthy that $\alpha^2=0$ corresponds to the linear or identity activation function, while $\alpha^2=1$ offers a smooth approximation of ReLU. This analysis indicates that the optimal degree of nonlinearity lies between these two extremes. Interestingly, we observe similar trends in Figure \ref{fig:filament_alpha} for the parameterized Swish activation function. In this instance, the majority of optimized $\alpha$ values fall within the range from $0$ to $1$, where $\alpha=0$ corresponds to the linear activation function, and $\alpha=1$ resembles ReLU. Consequently, our findings for the filament selection data set reaffirm that the suitable degree of nonlinearity for the trainable Swish activation function lies between the linear and ReLU functions.

In the final experiment within this section, we explore the influence of neural network architecture on the accuracy trade-off between fixed and adaptive activation functions. Recall that we initially set the number of hidden layers to $1$, and up to this point, our focus has been on the scenario with $2$ hidden units, i.e., $N_h=2$, as illustrated in Figure \ref{fig:architecture}. The main motivation for this choice was the small sample size in many scientific and engineering applications, including additive manufacturing problems. Thus, our objective was to determine the optimal structure of activation functions for a constrained number of hidden units. To broaden this analysis in the context of the filament selection problem, we now explore varying values of hidden units, specifically $N_h\in\{2,4,6,8\}$.

\begin{figure}[ht]
    \centering
    \includegraphics[width=\textwidth]{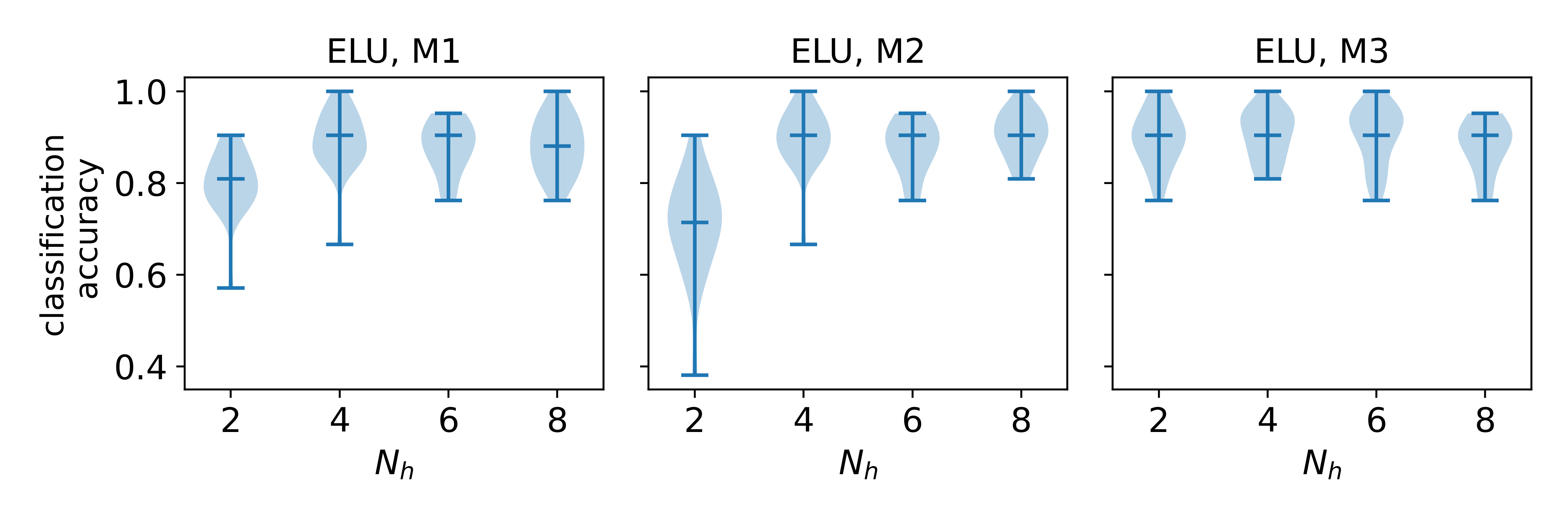}
    \caption{Investigating the impact of the number of hidden units $N_h$ on the performance of neural networks with fixed and adaptive ELU activation functions, shown in Figure \ref{fig:architecture}.}
    \label{fig:Nh}
\end{figure}

Figure \ref{fig:Nh} presents the classification accuracy scores achieved by ELU in M1, M2, and M3, with respect to the number of hidden units $N_h$. For the smallest neural network model with $N_h=2$, we observe that the adaptive ELU activation function with individual trainable parameters in M3 reaches the maximum accuracy score of $1$, outperforming both M1 and M2 by a significant margin. It is worth highlighting that the worst-case performance of M3 when $N_h=2$ is much more reasonable compared to M1 and M2. As we increase the number of hidden units to $N_h=4$, we see improvements in accuracy in the three models M1, M2, and M3. However, M3 still outperforms the other two models, especially in terms of the worst-case classification accuracy score.

Moreover, we observe that higher values of $N_h$ do not provide substantial accuracy improvements for both fixed and adaptive activation functions. As previously mentioned, this aligns with expectations due to the limited number of training points, which increases the risk of overfitting. This observation reinforces the rationale for employing adaptive activation functions with individual trainable parameters, as implemented in M3. Such an approach allows the training of small yet flexible and accurate neural networks, making them suitable for predictive modeling with sparse experimental data.

\section{Conclusion and Future Work}\label{sec:conc}
In this study, we investigated the adaptability of neural network models in scenarios with limited data, leveraging parameterized activation functions. Employing three real-world testbeds derived from additive manufacturing problems, we demonstrated that neural network models equipped with individual trainable parameters\textemdash going beyond the conventional practice of employing identical activation functions for a given hidden layer\textemdash resulted in improved prediction models. We evaluated these enhancements through the lens of point predictions, gauged by the standard classification accuracy score, and further delved into the realm of prediction sets through conformal inference. Specifically, we noted the following key observations.
\begin{enumerate}
    \item When dealing with sparse scientific data sets, opting for ELU and Softplus activation functions with individual trainable parameters proved advantageous over fixed and parameterized activation functions shared across all units in a hidden layer. The adoption of individual trainable activation functions in M3 demonstrated remarkable flexibility, allowing each unit to discern the optimal degree of nonlinearity introduced when conveying information to the next layer.
    \item Leveraging conformal inference emerged as a versatile and crucial approach to assess the confidence in the predictions of neural network models with trainable activation functions. Due to the increase in the number of parameters that must be inferred during the training process, measuring the empirical coverage score and the average prediction set size were informative measures to quantify predictive uncertainty. 
    \item The performance trade-offs between fixed and adaptive activation functions were heavily contingent on the neural network architecture. Consequently, in scenarios with limited data, striking the right balance between the number of hidden units and their adaptability becomes crucial.
    \item As automated machine learning (AutoML) methods gain popularity to democratize the utilization of prediction models among practitioners and engineers \cite{jin2023autokeras}, incorporating adaptive activation functions can yield substantial improvements. This approach mitigates the dependence on predetermined and fixed-shape activation functions, which can significantly impact predictive performance.
\end{enumerate}

In future work, we plan to explore several extensions of our study. In particular, we aim to investigate the performance of adaptive activation functions in more complex neural network models, such as convolutional neural networks (CNNs). This exploration may find application in the development of adaptive CNNs for constructing accurate prediction models for in situ monitoring of manufacturing technologies with minimal reliance on human supervision. Another extension will revolve around the incorporation of ensemble activation functions. Here, the objective is to leverage richer parameterized activation functions, offering greater flexibility compared to the activation functions employed in this study.

\section*{Acknowledgment}
Research was sponsored by DEVCOM Army Research Laboratory and was accomplished under Cooperative Agreement Number W911NF-19-2-0100. The views and conclusions contained in this document are those of the authors and should not be interpreted as representing the official policies, either expressed or implied, of DEVCOM Army Research Laboratory or the U.S. Government. The U.S. Government is authorized to reproduce and
distribute reprints for Government purposes notwithstanding any copyright notation herein.

\section*{Conflicts of Interest}
The authors declare that they have no known competing financial interests or personal relationships that could have appeared to influence the work reported in this paper.

% \end{appendices}

%%===========================================================================================%%
%% If you are submitting to one of the Nature Portfolio journals, using the eJP submission   %%
%% system, please include the references within the manuscript file itself. You may do this  %%
%% by copying the reference list from your .bbl file, paste it into the main manuscript .tex %%
%% file, and delete the associated \verb+\bibliography+ commands.                            %%
%%===========================================================================================%%

\bibliography{sn-bibliography}% common bib file

%% BioMed_Central_Bib_Style_v1.01

\begin{thebibliography}{53}
% BibTex style file: bmc-mathphys.bst (version 2.1), 2014-07-24
\ifx \bisbn   \undefined \def \bisbn  #1{ISBN #1}\fi
\ifx \binits  \undefined \def \binits#1{#1}\fi
\ifx \bauthor  \undefined \def \bauthor#1{#1}\fi
\ifx \batitle  \undefined \def \batitle#1{#1}\fi
\ifx \bjtitle  \undefined \def \bjtitle#1{#1}\fi
\ifx \bvolume  \undefined \def \bvolume#1{\textbf{#1}}\fi
\ifx \byear  \undefined \def \byear#1{#1}\fi
\ifx \bissue  \undefined \def \bissue#1{#1}\fi
\ifx \bfpage  \undefined \def \bfpage#1{#1}\fi
\ifx \blpage  \undefined \def \blpage #1{#1}\fi
\ifx \burl  \undefined \def \burl#1{\textsf{#1}}\fi
\ifx \doiurl  \undefined \def \doiurl#1{\url{https://doi.org/#1}}\fi
\ifx \betal  \undefined \def \betal{\textit{et al.}}\fi
\ifx \binstitute  \undefined \def \binstitute#1{#1}\fi
\ifx \binstitutionaled  \undefined \def \binstitutionaled#1{#1}\fi
\ifx \bctitle  \undefined \def \bctitle#1{#1}\fi
\ifx \beditor  \undefined \def \beditor#1{#1}\fi
\ifx \bpublisher  \undefined \def \bpublisher#1{#1}\fi
\ifx \bbtitle  \undefined \def \bbtitle#1{#1}\fi
\ifx \bedition  \undefined \def \bedition#1{#1}\fi
\ifx \bseriesno  \undefined \def \bseriesno#1{#1}\fi
\ifx \blocation  \undefined \def \blocation#1{#1}\fi
\ifx \bsertitle  \undefined \def \bsertitle#1{#1}\fi
\ifx \bsnm \undefined \def \bsnm#1{#1}\fi
\ifx \bsuffix \undefined \def \bsuffix#1{#1}\fi
\ifx \bparticle \undefined \def \bparticle#1{#1}\fi
\ifx \barticle \undefined \def \barticle#1{#1}\fi
\bibcommenthead
\ifx \bconfdate \undefined \def \bconfdate #1{#1}\fi
\ifx \botherref \undefined \def \botherref #1{#1}\fi
\ifx \url \undefined \def \url#1{\textsf{#1}}\fi
\ifx \bchapter \undefined \def \bchapter#1{#1}\fi
\ifx \bbook \undefined \def \bbook#1{#1}\fi
\ifx \bcomment \undefined \def \bcomment#1{#1}\fi
\ifx \oauthor \undefined \def \oauthor#1{#1}\fi
\ifx \citeauthoryear \undefined \def \citeauthoryear#1{#1}\fi
\ifx \endbibitem  \undefined \def \endbibitem {}\fi
\ifx \bconflocation  \undefined \def \bconflocation#1{#1}\fi
\ifx \arxivurl  \undefined \def \arxivurl#1{\textsf{#1}}\fi
\csname PreBibitemsHook\endcsname

%%% 1
\bibitem[\protect\citeauthoryear{Lu and Lu}{2020}]{lu2020universal}
\begin{barticle}
\bauthor{\bsnm{Lu}, \binits{Y.}},
\bauthor{\bsnm{Lu}, \binits{J.}}:
\batitle{A universal approximation theorem of deep neural networks for
  expressing probability distributions}.
\bjtitle{Advances in Neural Information Processing Systems}
\bvolume{33},
\bfpage{3094}--\blpage{3105}
(\byear{2020})
\end{barticle}
\endbibitem

%%% 2
\bibitem[\protect\citeauthoryear{Talaei~Khoei et~al.}{2023}]{talaei2023deep}
\begin{botherref}
\oauthor{\bsnm{Talaei~Khoei}, \binits{T.}},
\oauthor{\bsnm{Ould~Slimane}, \binits{H.}},
\oauthor{\bsnm{Kaabouch}, \binits{N.}}:
Deep learning: systematic review, models, challenges, and research directions.
Neural Computing and Applications,
1--22
(2023)
\end{botherref}
\endbibitem

%%% 3
\bibitem[\protect\citeauthoryear{Abdou}{2022}]{abdou2022literature}
\begin{barticle}
\bauthor{\bsnm{Abdou}, \binits{M.}}:
\batitle{Literature review: Efficient deep neural networks techniques for
  medical image analysis}.
\bjtitle{Neural Computing and Applications}
\bvolume{34}(\bissue{8}),
\bfpage{5791}--\blpage{5812}
(\byear{2022})
\end{barticle}
\endbibitem

%%% 4
\bibitem[\protect\citeauthoryear{Weiss et~al.}{2022}]{weiss2022applications}
\begin{barticle}
\bauthor{\bsnm{Weiss}, \binits{R.}},
\bauthor{\bsnm{Karimijafarbigloo}, \binits{S.}},
\bauthor{\bsnm{Roggenbuck}, \binits{D.}},
\bauthor{\bsnm{R{\"o}diger}, \binits{S.}}:
\batitle{Applications of neural networks in biomedical data analysis}.
\bjtitle{Biomedicines}
\bvolume{10}(\bissue{7}),
\bfpage{1469}
(\byear{2022})
\end{barticle}
\endbibitem

%%% 5
\bibitem[\protect\citeauthoryear{Liu et~al.}{2023}]{liu2023development}
\begin{barticle}
\bauthor{\bsnm{Liu}, \binits{X.}},
\bauthor{\bsnm{Miramini}, \binits{S.}},
\bauthor{\bsnm{Patel}, \binits{M.}},
\bauthor{\bsnm{Ebeling}, \binits{P.}},
\bauthor{\bsnm{Liao}, \binits{J.}},
\bauthor{\bsnm{Zhang}, \binits{L.}}:
\batitle{Development of numerical model-based machine learning algorithms for
  different healing stages of distal radius fracture healing}.
\bjtitle{Computer Methods and Programs in Biomedicine}
\bvolume{233},
\bfpage{107464}
(\byear{2023})
\end{barticle}
\endbibitem

%%% 6
\bibitem[\protect\citeauthoryear{Pourkamali-Anaraki and
  Hariri-Ardebili}{2021}]{pourkamali2021neural}
\begin{barticle}
\bauthor{\bsnm{Pourkamali-Anaraki}, \binits{F.}},
\bauthor{\bsnm{Hariri-Ardebili}, \binits{M.}}:
\batitle{Neural networks and imbalanced learning for data-driven scientific
  computing with uncertainties}.
\bjtitle{IEEE Access}
\bvolume{9},
\bfpage{15334}--\blpage{15350}
(\byear{2021})
\end{barticle}
\endbibitem

%%% 7
\bibitem[\protect\citeauthoryear{Khodadadi~Koodiani
  et~al.}{2023}]{khodadadi2023non}
\begin{barticle}
\bauthor{\bsnm{Khodadadi~Koodiani}, \binits{H.}},
\bauthor{\bsnm{Majlesi}, \binits{A.}},
\bauthor{\bsnm{Shahriar}, \binits{A.}},
\bauthor{\bsnm{Matamoros}, \binits{A.}}:
\batitle{Non-linear modeling parameters for new construction rc columns}.
\bjtitle{Frontiers in Built Environment}
\bvolume{9},
\bfpage{1108319}
(\byear{2023})
\end{barticle}
\endbibitem

%%% 8
\bibitem[\protect\citeauthoryear{Olivier et~al.}{2021}]{olivier2021bayesian}
\begin{barticle}
\bauthor{\bsnm{Olivier}, \binits{A.}},
\bauthor{\bsnm{Shields}, \binits{M.}},
\bauthor{\bsnm{Graham-Brady}, \binits{L.}}:
\batitle{Bayesian neural networks for uncertainty quantification in data-driven
  materials modeling}.
\bjtitle{Computer methods in applied mechanics and engineering}
\bvolume{386},
\bfpage{114079}
(\byear{2021})
\end{barticle}
\endbibitem

%%% 9
\bibitem[\protect\citeauthoryear{Stuckner et~al.}{2021}]{stuckner2021optimal}
\begin{barticle}
\bauthor{\bsnm{Stuckner}, \binits{J.}},
\bauthor{\bsnm{Piekenbrock}, \binits{M.}},
\bauthor{\bsnm{Arnold}, \binits{S.}},
\bauthor{\bsnm{Ricks}, \binits{T.}}:
\batitle{Optimal experimental design with fast neural network surrogate
  models}.
\bjtitle{Computational Materials Science}
\bvolume{200},
\bfpage{110747}
(\byear{2021})
\end{barticle}
\endbibitem

%%% 10
\bibitem[\protect\citeauthoryear{Brunton et~al.}{2020}]{brunton2020special}
\begin{barticle}
\bauthor{\bsnm{Brunton}, \binits{S.}},
\bauthor{\bsnm{Hemati}, \binits{M.}},
\bauthor{\bsnm{Taira}, \binits{K.}}:
\batitle{Special issue on machine learning and data-driven methods in fluid
  dynamics}.
\bjtitle{Theoretical and Computational Fluid Dynamics}
\bvolume{34}(\bissue{4}),
\bfpage{333}--\blpage{337}
(\byear{2020})
\end{barticle}
\endbibitem

%%% 11
\bibitem[\protect\citeauthoryear{Erichson et~al.}{2020}]{erichson2020shallow}
\begin{barticle}
\bauthor{\bsnm{Erichson}, \binits{B.}},
\bauthor{\bsnm{Mathelin}, \binits{L.}},
\bauthor{\bsnm{Yao}, \binits{Z.}},
\bauthor{\bsnm{Brunton}, \binits{S.}},
\bauthor{\bsnm{Mahoney}, \binits{M.}},
\bauthor{\bsnm{Kutz}, \binits{N.}}:
\batitle{Shallow neural networks for fluid flow reconstruction with limited
  sensors}.
\bjtitle{Proceedings of the Royal Society A}
\bvolume{476}(\bissue{2238}),
\bfpage{20200097}
(\byear{2020})
\end{barticle}
\endbibitem

%%% 12
\bibitem[\protect\citeauthoryear{Johnson et~al.}{2020}]{johnson2020invited}
\begin{barticle}
\bauthor{\bsnm{Johnson}, \binits{N.}},
\bauthor{\bsnm{Vulimiri}, \binits{P.}},
\bauthor{\bsnm{To}, \binits{A.}},
\bauthor{\bsnm{Zhang}, \binits{X.}},
\bauthor{\bsnm{Brice}, \binits{C.}},
\bauthor{\bsnm{Kappes}, \binits{B.}},
\bauthor{\bsnm{Stebner}, \binits{A.}}:
\batitle{Invited review: Machine learning for materials developments in metals
  additive manufacturing}.
\bjtitle{Additive Manufacturing}
\bvolume{36},
\bfpage{101641}
(\byear{2020})
\end{barticle}
\endbibitem

%%% 13
\bibitem[\protect\citeauthoryear{Pourkamali-Anaraki
  et~al.}{2023}]{pourkamali2023evaluation}
\begin{barticle}
\bauthor{\bsnm{Pourkamali-Anaraki}, \binits{F.}},
\bauthor{\bsnm{Nasrin}, \binits{T.}},
\bauthor{\bsnm{Jensen}, \binits{R.}},
\bauthor{\bsnm{Peterson}, \binits{A.}},
\bauthor{\bsnm{Hansen}, \binits{C.}}:
\batitle{Evaluation of classification models in limited data scenarios with
  application to additive manufacturing}.
\bjtitle{Engineering Applications of Artificial Intelligence}
\bvolume{126},
\bfpage{106983}
(\byear{2023})
\end{barticle}
\endbibitem

%%% 14
\bibitem[\protect\citeauthoryear{Hayou et~al.}{2019}]{hayou2019impact}
\begin{bchapter}
\bauthor{\bsnm{Hayou}, \binits{S.}},
\bauthor{\bsnm{Doucet}, \binits{A.}},
\bauthor{\bsnm{Rousseau}, \binits{J.}}:
\bctitle{On the impact of the activation function on deep neural networks
  training}.
In: \bbtitle{International Conference on Machine Learning},
pp. \bfpage{2672}--\blpage{2680}
(\byear{2019})
\end{bchapter}
\endbibitem

%%% 15
\bibitem[\protect\citeauthoryear{Hu et~al.}{2021}]{hu2021handling}
\begin{barticle}
\bauthor{\bsnm{Hu}, \binits{Z.}},
\bauthor{\bsnm{Zhang}, \binits{J.}},
\bauthor{\bsnm{Ge}, \binits{Y.}}:
\batitle{Handling vanishing gradient problem using artificial derivative}.
\bjtitle{IEEE Access}
\bvolume{9},
\bfpage{22371}--\blpage{22377}
(\byear{2021})
\end{barticle}
\endbibitem

%%% 16
\bibitem[\protect\citeauthoryear{Shen et~al.}{2022}]{shen2022enhancement}
\begin{barticle}
\bauthor{\bsnm{Shen}, \binits{S.}},
\bauthor{\bsnm{Zhang}, \binits{N.}},
\bauthor{\bsnm{Zhou}, \binits{A.}},
\bauthor{\bsnm{Yin}, \binits{Z.}}:
\batitle{Enhancement of neural networks with an alternative activation function
  tanhlu}.
\bjtitle{Expert Systems with Applications}
\bvolume{199},
\bfpage{117181}
(\byear{2022})
\end{barticle}
\endbibitem

%%% 17
\bibitem[\protect\citeauthoryear{Clevert et~al.}{2015}]{clevert2015fast}
\begin{botherref}
\oauthor{\bsnm{Clevert}, \binits{D.}},
\oauthor{\bsnm{Unterthiner}, \binits{T.}},
\oauthor{\bsnm{Hochreiter}, \binits{S.}}:
Fast and accurate deep network learning by exponential linear units (elus).
arXiv preprint arXiv:1511.07289
(2015)
\end{botherref}
\endbibitem

%%% 18
\bibitem[\protect\citeauthoryear{Zheng et~al.}{2015}]{zheng2015improving}
\begin{bchapter}
\bauthor{\bsnm{Zheng}, \binits{H.}},
\bauthor{\bsnm{Yang}, \binits{Z.}},
\bauthor{\bsnm{Liu}, \binits{W.}},
\bauthor{\bsnm{Liang}, \binits{J.}},
\bauthor{\bsnm{Li}, \binits{Y.}}:
\bctitle{Improving deep neural networks using softplus units}.
In: \bbtitle{International Joint Conference on Neural Networks},
pp. \bfpage{1}--\blpage{4}
(\byear{2015})
\end{bchapter}
\endbibitem

%%% 19
\bibitem[\protect\citeauthoryear{Ramachandran
  et~al.}{2017}]{ramachandran2017searching}
\begin{botherref}
\oauthor{\bsnm{Ramachandran}, \binits{P.}},
\oauthor{\bsnm{Zoph}, \binits{B.}},
\oauthor{\bsnm{Le}, \binits{Q.}}:
Searching for activation functions.
arXiv preprint arXiv:1710.05941
(2017)
\end{botherref}
\endbibitem

%%% 20
\bibitem[\protect\citeauthoryear{Chollet}{2021}]{chollet2021deep}
\begin{bbook}
\bauthor{\bsnm{Chollet}, \binits{F.}}:
\bbtitle{Deep Learning with {P}ython}.
\bpublisher{Simon and Schuster}, \blocation{???}
(\byear{2021})
\end{bbook}
\endbibitem

%%% 21
\bibitem[\protect\citeauthoryear{Agostinelli
  et~al.}{2014}]{agostinelli2014learning}
\begin{botherref}
\oauthor{\bsnm{Agostinelli}, \binits{F.}},
\oauthor{\bsnm{Hoffman}, \binits{M.}},
\oauthor{\bsnm{Sadowski}, \binits{P.}},
\oauthor{\bsnm{Baldi}, \binits{P.}}:
Learning activation functions to improve deep neural networks.
arXiv preprint arXiv:1412.6830
(2014)
\end{botherref}
\endbibitem

%%% 22
\bibitem[\protect\citeauthoryear{Lee et~al.}{2022}]{lee2022stochastic}
\begin{botherref}
\oauthor{\bsnm{Lee}, \binits{K.}},
\oauthor{\bsnm{Yang}, \binits{J.}},
\oauthor{\bsnm{Lee}, \binits{H.}},
\oauthor{\bsnm{Hwang}, \binits{J.}}:
Stochastic adaptive activation function.
Advances in Neural Information Processing Systems,
13787--13799
(2022)
\end{botherref}
\endbibitem

%%% 23
\bibitem[\protect\citeauthoryear{Dubey et~al.}{2022}]{dubey2022activation}
\begin{botherref}
\oauthor{\bsnm{Dubey}, \binits{S.}},
\oauthor{\bsnm{Singh}, \binits{S.}},
\oauthor{\bsnm{Chaudhuri}, \binits{B.}}:
Activation functions in deep learning: A comprehensive survey and benchmark.
Neurocomputing
(2022)
\end{botherref}
\endbibitem

%%% 24
\bibitem[\protect\citeauthoryear{Apicella et~al.}{2021}]{apicella2021survey}
\begin{barticle}
\bauthor{\bsnm{Apicella}, \binits{A.}},
\bauthor{\bsnm{Donnarumma}, \binits{F.}},
\bauthor{\bsnm{Isgr{\`o}}, \binits{F.}},
\bauthor{\bsnm{Prevete}, \binits{R.}}:
\batitle{A survey on modern trainable activation functions}.
\bjtitle{Neural Networks}
\bvolume{138},
\bfpage{14}--\blpage{32}
(\byear{2021})
\end{barticle}
\endbibitem

%%% 25
\bibitem[\protect\citeauthoryear{Shafer and Vovk}{2008}]{shafer2008tutorial}
\begin{barticle}
\bauthor{\bsnm{Shafer}, \binits{G.}},
\bauthor{\bsnm{Vovk}, \binits{V.}}:
\batitle{A tutorial on conformal prediction}.
\bjtitle{Journal of Machine Learning Research}
\bvolume{9}(\bissue{3}),
\bfpage{371}--\blpage{421}
(\byear{2008})
\end{barticle}
\endbibitem

%%% 26
\bibitem[\protect\citeauthoryear{Barber et~al.}{2023}]{barber2023conformal}
\begin{barticle}
\bauthor{\bsnm{Barber}, \binits{R.}},
\bauthor{\bsnm{Candes}, \binits{E.}},
\bauthor{\bsnm{Ramdas}, \binits{A.}},
\bauthor{\bsnm{Tibshirani}, \binits{R.}}:
\batitle{Conformal prediction beyond exchangeability}.
\bjtitle{The Annals of Statistics}
\bvolume{51}(\bissue{2}),
\bfpage{816}--\blpage{845}
(\byear{2023})
\end{barticle}
\endbibitem

%%% 27
\bibitem[\protect\citeauthoryear{Ke and Huang}{2020}]{ke2020quality}
\begin{barticle}
\bauthor{\bsnm{Ke}, \binits{K.}},
\bauthor{\bsnm{Huang}, \binits{M.}}:
\batitle{Quality prediction for injection molding by using a multilayer
  perceptron neural network}.
\bjtitle{Polymers}
\bvolume{12}(\bissue{8}),
\bfpage{1812}
(\byear{2020})
\end{barticle}
\endbibitem

%%% 28
\bibitem[\protect\citeauthoryear{Ren et~al.}{2020}]{ren2020balanced}
\begin{barticle}
\bauthor{\bsnm{Ren}, \binits{J.}},
\bauthor{\bsnm{Yu}, \binits{C.}},
\bauthor{\bsnm{Ma}, \binits{X.}},
\bauthor{\bsnm{Zhao}, \binits{H.}},
\bauthor{\bsnm{Yi}, \binits{S.}}:
\batitle{Balanced meta-softmax for long-tailed visual recognition}.
\bjtitle{Advances in Neural Information Processing Systems}
\bvolume{33},
\bfpage{4175}--\blpage{4186}
(\byear{2020})
\end{barticle}
\endbibitem

%%% 29
\bibitem[\protect\citeauthoryear{Yang et~al.}{2023}]{yang2023dprelu}
\begin{barticle}
\bauthor{\bsnm{Yang}, \binits{D.}},
\bauthor{\bsnm{Ngoc}, \binits{K.}},
\bauthor{\bsnm{Shin}, \binits{I.}},
\bauthor{\bsnm{Hwang}, \binits{M.}}:
\batitle{{DPReLU}: Dynamic parametric rectified linear unit and its proper
  weight initialization method}.
\bjtitle{International Journal of Computational Intelligence Systems}
\bvolume{16}(\bissue{1}),
\bfpage{11}
(\byear{2023})
\end{barticle}
\endbibitem

%%% 30
\bibitem[\protect\citeauthoryear{Zhu et~al.}{2021}]{zhu2021logish}
\begin{barticle}
\bauthor{\bsnm{Zhu}, \binits{H.}},
\bauthor{\bsnm{Zeng}, \binits{H.}},
\bauthor{\bsnm{Liu}, \binits{J.}},
\bauthor{\bsnm{Zhang}, \binits{X.}}:
\batitle{Logish: A new nonlinear nonmonotonic activation function for
  convolutional neural network}.
\bjtitle{Neurocomputing}
\bvolume{458},
\bfpage{490}--\blpage{499}
(\byear{2021})
\end{barticle}
\endbibitem

%%% 31
\bibitem[\protect\citeauthoryear{{\c{C}}atalba{\c{s}} and
  Morg{\"u}l}{2023}]{ccatalbacs2023deep}
\begin{botherref}
\oauthor{\bsnm{{\c{C}}atalba{\c{s}}}, \binits{B.}},
\oauthor{\bsnm{Morg{\"u}l}, \binits{{\"O}.}}:
Deep learning with {ExtendeD Exponential Linear Unit (DELU)}.
Neural Computing and Applications,
22705--22724
(2023)
\end{botherref}
\endbibitem

%%% 32
\bibitem[\protect\citeauthoryear{Emanuel et~al.}{2023}]{emanuel2023effect}
\begin{botherref}
\oauthor{\bsnm{Emanuel}, \binits{R.}},
\oauthor{\bsnm{Docherty}, \binits{P.}},
\oauthor{\bsnm{Lunt}, \binits{H.}},
\oauthor{\bsnm{M{\"o}ller}, \binits{K.}}:
The effect of activation functions on accuracy, convergence speed, and
  misclassification confidence in {CNN} text classification: a comprehensive
  exploration.
The Journal of Supercomputing,
1--21
(2023)
\end{botherref}
\endbibitem

%%% 33
\bibitem[\protect\citeauthoryear{Wang et~al.}{2022}]{wang2022kdac}
\begin{barticle}
\bauthor{\bsnm{Wang}, \binits{Z.}},
\bauthor{\bsnm{Liu}, \binits{H.}},
\bauthor{\bsnm{Liu}, \binits{F.}},
\bauthor{\bsnm{Gao}, \binits{D.}}:
\batitle{Why {KDAC}? {A} general activation function for knowledge discovery}.
\bjtitle{Neurocomputing}
\bvolume{501},
\bfpage{343}--\blpage{358}
(\byear{2022})
\end{barticle}
\endbibitem

%%% 34
\bibitem[\protect\citeauthoryear{Klopries and
  Schwung}{2023}]{klopries2023flexible}
\begin{bchapter}
\bauthor{\bsnm{Klopries}, \binits{H.}},
\bauthor{\bsnm{Schwung}, \binits{A.}}:
\bctitle{Flexible activation bag: Learning activation functions in autoencoder
  networks}.
In: \bbtitle{IEEE International Conference on Industrial Technology (ICIT)},
pp. \bfpage{1}--\blpage{7}
(\byear{2023})
\end{bchapter}
\endbibitem

%%% 35
\bibitem[\protect\citeauthoryear{Jagtap and
  Karniadakis}{2023}]{jagtap2023important}
\begin{botherref}
\oauthor{\bsnm{Jagtap}, \binits{A.}},
\oauthor{\bsnm{Karniadakis}, \binits{G.}}:
How important are activation functions in regression and classification? a
  survey, performance comparison, and future directions.
Journal of Machine Learning for Modeling and Computing
\textbf{4}(1)
(2023)
\end{botherref}
\endbibitem

%%% 36
\bibitem[\protect\citeauthoryear{Gnanasambandam et~al.}{2023}]{10227556}
\begin{barticle}
\bauthor{\bsnm{Gnanasambandam}, \binits{R.}},
\bauthor{\bsnm{Shen}, \binits{B.}},
\bauthor{\bsnm{Chung}, \binits{J.}},
\bauthor{\bsnm{Yue}, \binits{X.}},
\bauthor{\bsnm{Kong}, \binits{Z.}}:
\batitle{Self-scalable {Tanh} {(Stan)}: Multi-scale solutions for
  physics-informed neural networks}.
\bjtitle{IEEE Transactions on Pattern Analysis and Machine Intelligence}
\bvolume{45}(\bissue{12}),
\bfpage{15588}--\blpage{15603}
(\byear{2023})
\end{barticle}
\endbibitem

%%% 37
\bibitem[\protect\citeauthoryear{Angelopoulos and
  Bates}{2023}]{angelopoulos2023conformal}
\begin{barticle}
\bauthor{\bsnm{Angelopoulos}, \binits{A.}},
\bauthor{\bsnm{Bates}, \binits{S.}}:
\batitle{Conformal prediction: A gentle introduction}.
\bjtitle{Foundations and Trends in Machine Learning}
\bvolume{16}(\bissue{4}),
\bfpage{494}--\blpage{591}
(\byear{2023})
\end{barticle}
\endbibitem

%%% 38
\bibitem[\protect\citeauthoryear{Lee et~al.}{2019}]{lee2019fabrication}
\begin{barticle}
\bauthor{\bsnm{Lee}, \binits{J.}},
\bauthor{\bsnm{Lee}, \binits{H.}},
\bauthor{\bsnm{Cheon}, \binits{K.}},
\bauthor{\bsnm{Park}, \binits{C.}},
\bauthor{\bsnm{Jang}, \binits{T.}},
\bauthor{\bsnm{Kim}, \binits{H.}},
\bauthor{\bsnm{Jung}, \binits{H.}}:
\batitle{Fabrication of poly (lactic acid)/{Ti} composite scaffolds with
  enhanced mechanical properties and biocompatibility via fused filament
  fabrication ({FFF})--based {3D} printing}.
\bjtitle{Additive Manufacturing}
\bvolume{30},
\bfpage{100883}
(\byear{2019})
\end{barticle}
\endbibitem

%%% 39
\bibitem[\protect\citeauthoryear{Wu et~al.}{2018}]{WU2018298}
\begin{barticle}
\bauthor{\bsnm{Wu}, \binits{H.}},
\bauthor{\bsnm{Sulkis}, \binits{M.}},
\bauthor{\bsnm{Driver}, \binits{J.}},
\bauthor{\bsnm{Saade-Castillo}, \binits{A.}},
\bauthor{\bsnm{Thompson}, \binits{A.}},
\bauthor{\bsnm{Koo}, \binits{J.}}:
\batitle{Multi-functional {ULTEM1010} composite filaments for additive
  manufacturing using fused filament fabrication ({FFF})}.
\bjtitle{Additive Manufacturing}
\bvolume{24},
\bfpage{298}--\blpage{306}
(\byear{2018})
\end{barticle}
\endbibitem

%%% 40
\bibitem[\protect\citeauthoryear{Pei et~al.}{2022}]{pei2022combining}
\begin{barticle}
\bauthor{\bsnm{Pei}, \binits{H.}},
\bauthor{\bsnm{Shi}, \binits{S.}},
\bauthor{\bsnm{Chen}, \binits{Y.}},
\bauthor{\bsnm{Xiong}, \binits{Y.}},
\bauthor{\bsnm{Lv}, \binits{Q.}}:
\batitle{Combining solid-state shear milling and {FFF} {3D}-printing strategy
  to fabricate high-performance biomimetic wearable fish-scale {PVDF}-based
  piezoelectric energy harvesters}.
\bjtitle{ACS Applied Materials \& Interfaces}
\bvolume{14}(\bissue{13}),
\bfpage{15346}--\blpage{15359}
(\byear{2022})
\end{barticle}
\endbibitem

%%% 41
\bibitem[\protect\citeauthoryear{Goh et~al.}{2020}]{goh2020process}
\begin{barticle}
\bauthor{\bsnm{Goh}, \binits{G.}},
\bauthor{\bsnm{Yap}, \binits{Y.}},
\bauthor{\bsnm{Tan}, \binits{H.}},
\bauthor{\bsnm{Sing}, \binits{S.}},
\bauthor{\bsnm{Goh}, \binits{G.}},
\bauthor{\bsnm{Yeong}, \binits{W.}}:
\batitle{Process--structure--properties in polymer additive manufacturing via
  material extrusion: A review}.
\bjtitle{Critical Reviews in Solid State and Materials Sciences}
\bvolume{45}(\bissue{2}),
\bfpage{113}--\blpage{133}
(\byear{2020})
\end{barticle}
\endbibitem

%%% 42
\bibitem[\protect\citeauthoryear{}{2023}]{apmonitor}
\begin{botherref}
Additive Manufacturing.
\url{https://apmonitor.com/pds/index.php/Main/AdditiveManufacturing}
\end{botherref}
\endbibitem

%%% 43
\bibitem[\protect\citeauthoryear{Braconnier
  et~al.}{2020}]{braconnier2020processing}
\begin{barticle}
\bauthor{\bsnm{Braconnier}, \binits{D.}},
\bauthor{\bsnm{Jensen}, \binits{R.}},
\bauthor{\bsnm{Peterson}, \binits{A.}}:
\batitle{Processing parameter correlations in material extrusion additive
  manufacturing}.
\bjtitle{Additive Manufacturing}
\bvolume{31},
\bfpage{100924}
(\byear{2020})
\end{barticle}
\endbibitem

%%% 44
\bibitem[\protect\citeauthoryear{Gao et~al.}{2021}]{gao2021fused}
\begin{barticle}
\bauthor{\bsnm{Gao}, \binits{X.}},
\bauthor{\bsnm{Qi}, \binits{S.}},
\bauthor{\bsnm{Kuang}, \binits{X.}},
\bauthor{\bsnm{Su}, \binits{Y.}},
\bauthor{\bsnm{Li}, \binits{J.}},
\bauthor{\bsnm{Wang}, \binits{D.}}:
\batitle{Fused filament fabrication of polymer materials: A review of
  interlayer bond}.
\bjtitle{Additive Manufacturing}
\bvolume{37},
\bfpage{101658}
(\byear{2021})
\end{barticle}
\endbibitem

%%% 45
\bibitem[\protect\citeauthoryear{Shah et~al.}{2021}]{shah2021highly}
\begin{barticle}
\bauthor{\bsnm{Shah}, \binits{D.}},
\bauthor{\bsnm{Morris}, \binits{J.}},
\bauthor{\bsnm{Plaisted}, \binits{T.}},
\bauthor{\bsnm{Amirkhizi}, \binits{A.}},
\bauthor{\bsnm{Hansen}, \binits{C.}}:
\batitle{Highly filled resins for {DLP}-based printing of low density, high
  modulus materials}.
\bjtitle{Additive Manufacturing}
\bvolume{37},
\bfpage{101736}
(\byear{2021})
\end{barticle}
\endbibitem

%%% 46
\bibitem[\protect\citeauthoryear{Zakeri et~al.}{2020}]{zakeri2020comprehensive}
\begin{barticle}
\bauthor{\bsnm{Zakeri}, \binits{S.}},
\bauthor{\bsnm{Vippola}, \binits{M.}},
\bauthor{\bsnm{Lev{\"a}nen}, \binits{E.}}:
\batitle{A comprehensive review of the photopolymerization of ceramic resins
  used in stereolithography}.
\bjtitle{Additive Manufacturing}
\bvolume{35},
\bfpage{101177}
(\byear{2020})
\end{barticle}
\endbibitem

%%% 47
\bibitem[\protect\citeauthoryear{Wang et~al.}{2020}]{wang2020fabrication}
\begin{barticle}
\bauthor{\bsnm{Wang}, \binits{W.}},
\bauthor{\bsnm{Sun}, \binits{J.}},
\bauthor{\bsnm{Guo}, \binits{B.}},
\bauthor{\bsnm{Chen}, \binits{X.}},
\bauthor{\bsnm{Ananth}, \binits{K.}},
\bauthor{\bsnm{Bai}, \binits{J.}}:
\batitle{Fabrication of piezoelectric nano-ceramics via stereolithography of
  low viscous and non-aqueous suspensions}.
\bjtitle{Journal of the European Ceramic Society}
\bvolume{40}(\bissue{3}),
\bfpage{682}--\blpage{688}
(\byear{2020})
\end{barticle}
\endbibitem

%%% 48
\bibitem[\protect\citeauthoryear{Al~Rashid et~al.}{2021}]{al2021vat}
\begin{barticle}
\bauthor{\bsnm{Al~Rashid}, \binits{A.}},
\bauthor{\bsnm{Ahmed}, \binits{W.}},
\bauthor{\bsnm{Khalid}, \binits{M.}},
\bauthor{\bsnm{Koc}, \binits{M.}}:
\batitle{Vat photopolymerization of polymers and polymer composites: Processes
  and applications}.
\bjtitle{Additive Manufacturing}
\bvolume{47},
\bfpage{102279}
(\byear{2021})
\end{barticle}
\endbibitem

%%% 49
\bibitem[\protect\citeauthoryear{Konijn et~al.}{2014}]{konijn2014experimental}
\begin{barticle}
\bauthor{\bsnm{Konijn}, \binits{B.}},
\bauthor{\bsnm{Sanderink}, \binits{O.}},
\bauthor{\bsnm{Kruyt}, \binits{N.}}:
\batitle{Experimental study of the viscosity of suspensions: Effect of solid
  fraction, particle size and suspending liquid}.
\bjtitle{Powder technology}
\bvolume{266},
\bfpage{61}--\blpage{69}
(\byear{2014})
\end{barticle}
\endbibitem

%%% 50
\bibitem[\protect\citeauthoryear{Delarue et~al.}{2023}]{delarue2023increasing}
\begin{botherref}
\oauthor{\bsnm{Delarue}, \binits{A.}},
\oauthor{\bsnm{McAninch}, \binits{I.}},
\oauthor{\bsnm{Peterson}, \binits{A.}},
\oauthor{\bsnm{Hansen}, \binits{C.}}:
Increasing printable solid loading in digital light processing using a bimodal
  particle size distribution.
3D Printing and Additive Manufacturing
(2023)
\end{botherref}
\endbibitem

%%% 51
\bibitem[\protect\citeauthoryear{Tomeckova and
  Halloran}{2010a}]{tomeckova2010critical}
\begin{barticle}
\bauthor{\bsnm{Tomeckova}, \binits{V.}},
\bauthor{\bsnm{Halloran}, \binits{J.}}:
\batitle{Critical energy for photopolymerization of ceramic suspensions in
  acrylate monomers}.
\bjtitle{Journal of the European Ceramic Society}
\bvolume{30}(\bissue{16}),
\bfpage{3273}--\blpage{3282}
(\byear{2010})
\end{barticle}
\endbibitem

%%% 52
\bibitem[\protect\citeauthoryear{Tomeckova and
  Halloran}{2010b}]{tomeckova2010cure}
\begin{barticle}
\bauthor{\bsnm{Tomeckova}, \binits{V.}},
\bauthor{\bsnm{Halloran}, \binits{J.}}:
\batitle{Cure depth for photopolymerization of ceramic suspensions}.
\bjtitle{Journal of the European Ceramic Society}
\bvolume{30}(\bissue{15}),
\bfpage{3023}--\blpage{3033}
(\byear{2010})
\end{barticle}
\endbibitem

%%% 53
\bibitem[\protect\citeauthoryear{Jin et~al.}{2023}]{jin2023autokeras}
\begin{barticle}
\bauthor{\bsnm{Jin}, \binits{H.}},
\bauthor{\bsnm{Chollet}, \binits{F.}},
\bauthor{\bsnm{Song}, \binits{Q.}},
\bauthor{\bsnm{Hu}, \binits{X.}}:
\batitle{Autokeras: An {AutoML} library for deep learning}.
\bjtitle{Journal of Machine Learning Research}
\bvolume{24}(\bissue{6}),
\bfpage{1}--\blpage{6}
(\byear{2023})
\end{barticle}
\endbibitem

\end{thebibliography}
%% if required, the content of .bbl file can be included here once bbl is generated
%%\input sn-article.bbl

\end{document}